\pdfoutput=1

\documentclass[11pt,table,xcdraw]{article}

\usepackage[final]{coling}

\usepackage{fancyhdr}
\fancypagestyle{firstpage}{
  \fancyhf{}
  \fancyhead[L]{\normalsize Published in the Proceedings of the COLING 2025}
}

\usepackage{times}
\usepackage{latexsym}

\usepackage[T1]{fontenc}

\usepackage[utf8]{inputenc}

\usepackage{microtype}

\usepackage{inconsolata}

\usepackage{graphicx}

\usepackage{times}
\usepackage{epsfig}
\usepackage{amsmath}
\usepackage{amsmath}
\usepackage{amssymb}
\usepackage{grffile}

\usepackage{multirow}
\usepackage[normalem]{ulem}
\useunder{\uline}{\ul}{}

\usepackage{amsmath,scalerel}

\DeclareMathOperator*{\concat}{\scalerel*{\Vert}{\sum}}

\title{FS-DAG: Few Shot Domain Adapting Graph Networks for Visually Rich Document Understanding}

\author{
 \textbf{Amit Agarwal\textsuperscript{1}},
 \textbf{Srikant Panda\textsuperscript{2}},
 \textbf{Kulbhushan Pachuri\textsuperscript{2}}
\\
\\
 \textsuperscript{1} OCI, Oracle USA,
 \textsuperscript{2} OCI, Oracle India
\\
 \small{
   \textbf{Correspondence:} \href{mailto:email@domain}{amit.h.agarwal@oracle.com}
 }
}

\begin{document}
\thispagestyle{firstpage}
\pagestyle{firstpage}
\maketitle
\begin{abstract}

In this work, we propose Few Shot Domain Adapting Graph (FS-DAG), a scalable and efficient model architecture for visually rich document understanding (VRDU) in few-shot settings. FS-DAG leverages domain-specific and language/vision specific backbones within a modular framework to adapt to diverse document types with minimal data. The model is robust to practical challenges such as handling OCR errors, misspellings, and domain shifts, which are critical in real-world deployments. FS-DAG is highly performant with less than 90M parameters, making it well-suited for complex real-world applications for Information Extraction (IE) tasks where computational resources are limited. We demonstrate FS-DAG's capability through extensive experiments for information extraction task, showing significant improvements in convergence speed and performance compared to state-of-the-art methods. Additionally, this work highlights the ongoing progress in developing smaller, more efficient models that do not compromise on performance.

\end{abstract}

\section{Introduction}

Recent advancements of Vision-Language Models (VLMs) \citep{ zhang2024vision,pattnayak2024survey,agarwal2024mvtamperbench}, Large Multimodal Models (LMMs) \citep{chen2024internvl,li2024llava}, and Large Language Models (LLMs) \citep{brown2020language, touvron2023llama}, have significantly enhanced performance across various natural language processing \cite{pattnayak2025clinicalqa20multitask} and computer vision tasks. Despite their success, these models are often computationally expensive, requiring substantial resources that are impractical for many real-world industrial applications \citep{ sanh2019distilbert,kaddour2023challenges}. Furthermore, their ability to adapt to specific domains, especially in the context of visually rich documents (VRDs) remains limited due to the high cost of pre-training and fine-tuning on domain-specific data \citep{li2021structext}.

VRDs face challenges stemming from diverse layouts, domain-specific terminology, and text variations in style and size. OCR-free models tend to underperform compared to key-value models that utilize a separate OCR component, and even these models struggle with such variations. Large-scale models, with their monolithic architectures, often rely on vast data for domain adaptation, complicating their deployment. For example, state-of-the-art models like LayoutLM \citep{xu2020layoutlmv2} and its successors demand extensive fine-tuning for new domains, making their deployment both costly and time-consuming \citep{huang2022layoutlmv3}.

To address these issues, we introduce FS-DAG, a few-shot learning framework designed for domain-specific document understanding with less than 90M parameters. Few-shot learning methods have gained attention for their ability to train models with limited labeled data, which is crucial in industrial applications where data scarcity is a common challenge. Our approach leverages a modular architecture that integrates domain-specific and language-specific feature extractors, allowing FS-DAG to adapt quickly to new domains with minimal data, thereby overcoming the barriers associated with large-scale models \citep{lee2022formnet}.

Our approach emphasizes few-shot learning by leveraging Graph Neural Networks (GNNs) \citep{khemani2024review,wu2020comprehensive,yin2024continuous} to enable rapid adaptation, robustness to OCR errors, and reduced latency in real-world applications. We provide empirical evidence of the model's performance through extensive experiments, showing significant improvements over larger methods with more than 100M parameters. To summarize, we make the following contributions to VRDU in a few-shot learning environment:

1. A modular framework for few-shot learning that efficiently combines domain-specific and language-specific textual and visual feature extractors in a graph-based architecture.

2. We propose shared positional embedding \& consistent reading order for GNN along with various training strategies for the model's robustness and effective adaptation with minimal data.

3. We provide comprehensive experimental results demonstrating that FS-DAG achieves state-of-the-art performance and robustness in few-shot learning scenarios while reducing latency and computational costs.

\section{Related Work}

The development of efficient and scalable NLP models has gained significant attention in recent years, particularly with the rise of LLMs \citep{pattnayak9339review,pattnayak2025hybrid,patel2025sweeval} such as GPT-3 \citep{brown2020language}, LlaMa \citep{touvron2023llama}, Mixtrals \citep{jiang2024mixtral}. While these models have achieved remarkable success in various tasks, their application in industrial settings remains challenging due to their high computational demands and difficulty in adapting to domain-specific tasks.

Recent work have focused on enhancing the efficiency of these models through techniques such as model distillation \citep{sanh2019distilbert}, pruning \citep{cheng2024survey}, and efficient fine-tuning methods like LoRA \citep{hu2022lora,thomas2025model}. These approaches aim to reduce the computational cost of LLMs while maintaining their performance, making them more suitable for deployment in resource-constrained environments.

In the context of VRDU, graph-based models have shown promise, particularly in capturing the complex relationships between textual and visual elements in documents. Models such as SDMGR \citep{sun2021spatial}, DocParser \citep{rausch2021docparser}, PICK \citep{yu2021pick} and others \citep{liu2019graph,rastogi2020information,yao2021one} leverage GNNs to improve IE from documents. However, these models often require large amounts of training data and are not designed for quick adaptation to new domains.

FS-DAG builds on these approaches by introducing a few-shot learning framework that can efficiently adapt to new document types with minimal data. This capability is particularly important in industrial applications, where labeled data is often limited, and the ability to quickly adapt to new domains is crucial. Additionally, FS-DAG addresses practical challenges such as robustness to OCR errors and domain shifts, which are common in real-world deployments.

\begin{figure}[!th]
\begin{center} 
   \frame{\includegraphics[width=1\linewidth]{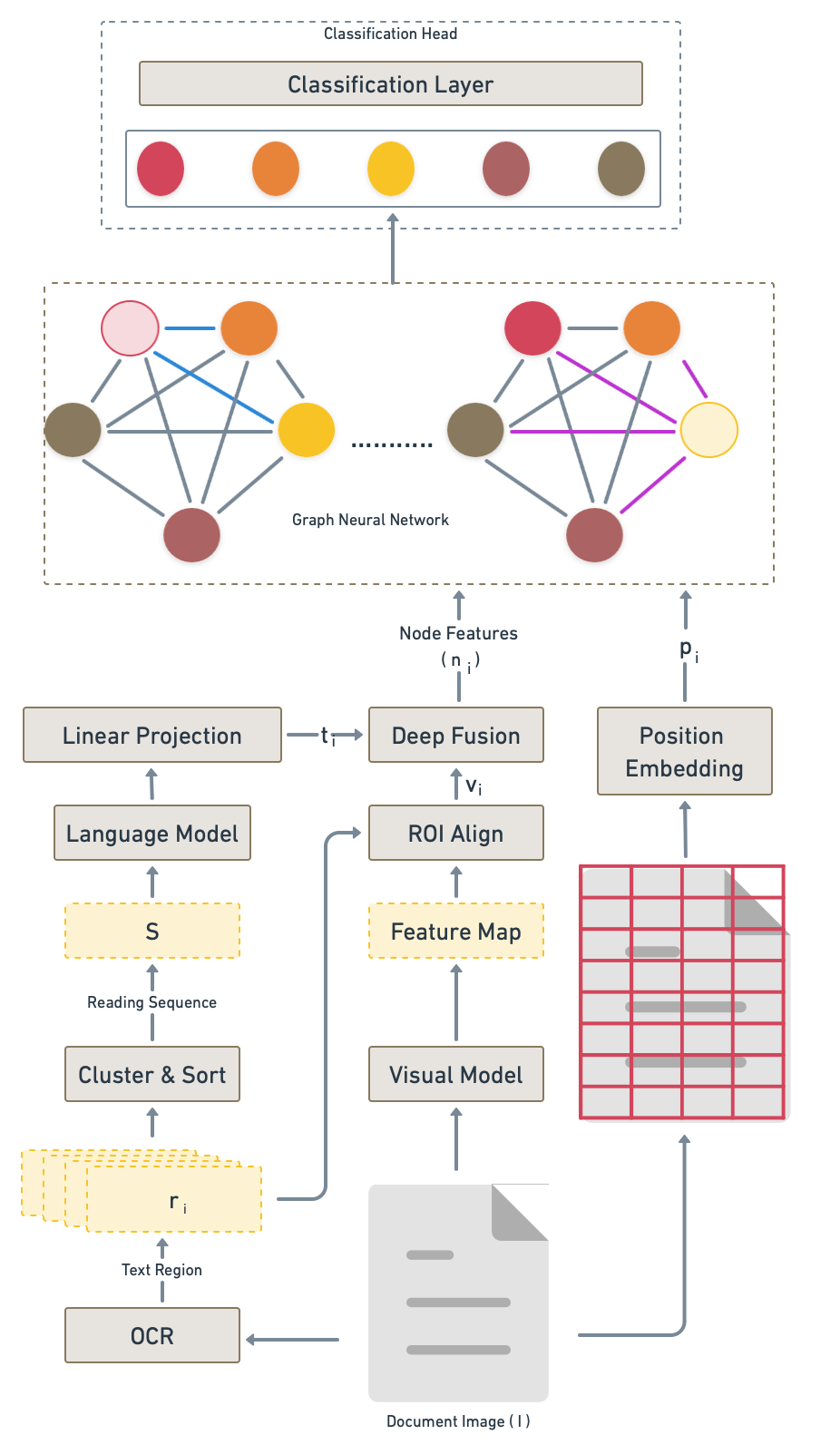}}
\end{center}
   \caption{An illustration of the model architecture for FS-DAG. Given a document image (I); its text regions $\{r_i\}$ are extracted using an OCR engine. We cluster and sort the $\{r_i\}$ to create a reading sequence $\{s\}$; textual features $\{t_i\}$ are extracted using a linear projection layer on top of a pre-trained language model processing $\{s\}$. In contrast, visual features $\{v_i\}$ are extracted using ROI-Align on top of the feature map from the Visual Model and $\{r_i\}$. The deep fusion module uses Kronecker product to fuse $\{t_i\}$ and $\{v_i\}$ to initialize the node features $\{n_i\}$. The node features are propagated and aggregated in the GNN during the message passing, which uses positional embedding $\{p_i\}$ and multi-head attention to learn the edge features dynamically. The classification head will finally classify the node features into one of the key-value classes.}
   \vspace{-1em}
\label{fig:onecol}
\end{figure}
\section{Our Approach}

Figure \ref{fig:onecol} illustrates our proposed model architecture. FS-DAG formulates the Key Information Extraction (KIE) \citep{huang2019icdar2019} task as a graph node classification problem using pre-trained feature extractors and graph multi-head attention in a few-shot learning environment. 

\subsection{Model Architecture}

The FS-DAG model \cite{agarwal2024domain} is designed to address the unique challenges associated with VRDU in few-shot learning scenarios. Unlike traditional monolithic models \citep{yu2021pick,xu2020layoutlm,xu2020layoutlmv2,huang2022layoutlmv3,xu2021layoutxlm} that often require large amounts of data and extensive computational resources, FS-DAG employs a modular architecture that efficiently integrates domain-specific and language-specific textual and visual feature extractors with a GNN. 

GNNs are particularly well-suited for VRDU tasks due to its ability to capture complex spatial and structural relationships between elements in a document. In FS-DAG, each document is represented as a graph where nodes correspond to these elements representing their textual and visual features, while the edges represent their spatial and semantic relationships. This graph representation allows the model to learn more robust and context-aware representations \cite{sun2021spatial, li2021structext}. FS-DAG further incorporates shared positional embeddings and a multi-head attention mechanism within the GNN. Shared positional embeddings provide a consistent reference for the spatial location of elements across different document types, while multi-head attention enables dynamic weighting of node connections, thereby improving feature aggregation and learning efficiency.

The FS-DAG architecture allows for the seamless integration of pre-trained domain-specific \cite{lee2020biobert, liu2021finbert} and language-specific feature extractors. This flexibility enables the model to quickly adapt to new domains with minimal data, significantly reducing the need for extensive retraining. By leveraging both textual and visual backbones tailored to specific domains, FS-DAG achieves superior performance compared to monolithic architectures that lack such adaptability. To further stabilize and boost the model's performance in a low-data setting, we modify the training strategies \cite{agarwal2023pseudo} and add augmentations for the graph \citep{agarwal2024techniques} and the visual modules. The individual components of the model are described further in the Appendix \ref{sec:appendix_model}.

\subsection{Training Strategies}

Training strategies are essential in few-shot training as we aim to attain the maximum model performance without overfitting the training dataset. To ensure higher performance and robustness of FS-DAG, we adopt various well-known strategies in the training process.

We include augmentation during training to enable the model to learn faster and be robust to various image and graph orientations. The augmentation technique focuses explicitly on the robustness of the visual embedding and the graph module. We introduce rotation ($\pm$ z degree), perspective transform, affine transform, and scaling and padding as the augmentations in the pipeline. These techniques enable the learning of better positional embeddings, visual embeddings, and node features as they change how the document is perceived and viewed. 
We also include specific graph augmentation \citep{agarwal2024techniques} which improves the convergence of FS-DAG with minimal data while making it robust to distribution shifts in textual or visual features

The proposed architecture does not support entity-linking currently and relies only on message propagation of the node features for the node classification task. Hence, 
we eliminate the edge loss function to stabilize the model training with the dedicated task. 

Owing to the inductive bias from the pre-trained feature extractors, we introduce Label Smoothing \citep{muller2019does} to the cross-entropy loss of node classification during training. Finally, to reduce overfitting in a few-shot learning paradigm, we add instance normalization \citep{ulyanov2016instance} 
over the node features of the graph. These changes enable us to train the model with better robustness and faster convergence. 

\section{Experiments}

FS-DAG is extensively evaluated on multiple datasets against state-of-the-art models based on their official implementations in terms of performance, robustness to OCR errors, and model complexity. The official open-source code base was used to compare the result with other state-of-the-art models, followed by hyper-parameter tuning to get the best results for a fair comparison.


All experiments were conducted thrice on a machine with 16 cores, 32GB of RAM. We trained FS-DAG using a node and edge embedding size of 64 and two GNN layers, with label smoothing set to 0.1. Due to the unavailability of official codebases for tasks, we could not benchmark architectures such as FormNet \citep{lee2022formnet} and StrucTexT \citep{li2021structext}. Few-shot techniques like LASER \citep{wang2022towards}, which do not leverage visual features, were also excluded from the comparison. Additionally, LMMs like LLaVA\citep{li2024llava}, Phi-3 \citep{abdin2024phi}, and InternVL \citep{chen2024internvl} were not benchmarked due to their considerable model size, which posed practical constraints. Other methods, such as \citep{orfew}, were omitted because they make multiple assumptions about the data structure and are not end-to-end trainable. To ensure a fair comparison, we focused on models with a size of less than 500M parameters.
\subsection{Datasets \& Metrics}

For the VRDU task of KIE, publicly available datasets such as SROIE \citep{huang2019icdar2019}, CORD \citep{park2019cord}, and WildReceipt \citep{sun2021spatial} primarily consist of document receipts from restaurants. While datasets like FUNSD \citep{jaume2019} and Kleister \citep{gralinski2020kleister} include various forms and longer documents, they typically focus on high-level key-value pairs. These datasets are valuable for academic research but often fall short of meeting the nuanced requirements of industry-specific data extraction, which demands handling fine-grained classes.

The majority of public datasets are concentrated on receipts, invoices, train tickets, and simple forms, which lack the diversity needed to cover the broad range of use cases in industry domains such as finance, healthcare, and logistics. These datasets also rarely capture documents that require detailed, character-by-character annotations within boxes or placeholders, which are highly relevant in industrial applications. Zilong \textit{et al.}\citep{wang2022benchmark} highlight these limitations and propose a new benchmark dataset for VRDU in both few-shot (10 and 50 samples) and conventional (100 and 200 samples settings. However, the document types in this dataset are limited to political ad-buys and registration forms, featuring only high-level fields ($\leq 10$) for extraction, thus not fully addressing the requirements of various industry verticals.

In this study, we use WildReceipt as a representative dataset from the existing public datasets, given its applicability to real-world receipt processing tasks. Additionally, we incorporate an industry-specific dataset\footnote{https://github.com/oracle-samples/fs-dag} 
that better reflects the characteristics needed across multiple domains, as outlined in Table \ref{tab:table_1}. This dataset includes document types filled character-by-character and features fine-grained key-value pair annotations at the word level, making it more aligned with the demands of industrial applications.
We compare state-of-the-art models under the same few-shot setting on these datasets and conduct an extensive ablation study on the proposed methods. Performance on the given datasets is evaluated using the F1 score, as defined by the ICDAR 2019 robust challenge \citep{huang2019icdar2019}, with the averaged F1 score over all classes being reported.


\begin{table}[t]
\begin{center}
\begin{tabular}{|c|l|c|}
\hline
\multicolumn{1}{|c|}{\begin{tabular}[c]{@{}c@{}}Dataset \\ Category\end{tabular}} &
\multicolumn{1}{c|}{\begin{tabular}[c]{@{}c@{}}Dataset \\ Name\end{tabular}} &
\begin{tabular}[c]{@{}c@{}}\# of \\ classes\end{tabular} \\ \hline

\multirow{5}{*}{1} 
& Ecommerce Invoice                                                  & 34   \\ \cline{2-3} 
& \begin{tabular}[c]{@{}l@{}}Adverse Reaction \\ Health Form \end{tabular} & 46   \\ \cline{2-3} 
& Medical Invoice                                                     & 33   \\ \cline{2-3} 
& \begin{tabular}[c]{@{}l@{}}University \\ Admission Form \end{tabular}   & 65   \\ \cline{2-3} 
& \begin{tabular}[c]{@{}l@{}}Visa Form \\ (Immigration) \end{tabular}     & 45   \\ \hline

\multirow{7}{*}{2} 
& Medical Authorization                                               & 34   \\ \cline{2-3} 
& \begin{tabular}[c]{@{}l@{}}Personal Bank \\ Account \end{tabular}       & 94   \\ \cline{2-3} 
& Equity Mortgage                                                     & 70   \\ \cline{2-3} 
& \begin{tabular}[c]{@{}l@{}}Corporate Bank \\ Account \end{tabular}      & 40   \\ \cline{2-3} 
& \begin{tabular}[c]{@{}l@{}}Online Banking \\ Application \end{tabular}  & 28   \\ \cline{2-3} 
& \begin{tabular}[c]{@{}l@{}}Medical Tax \\ Returns \end{tabular}         & 52   \\ \cline{2-3} 
& \begin{tabular}[c]{@{}l@{}}Medical Insurance \\ Enrollment \end{tabular} & 68   \\ \hline
\end{tabular}
\end{center}
\caption{Highlights the number of key-value classes across the each document type in the two  categories.}
\vspace{-1em}
\label{tab:table_1}
\end{table}

\begin{table*}[!th]
\small
\centering
\renewcommand{\arraystretch}{1.3} 
\setlength{\tabcolsep}{6pt} 
\begin{tabular}{|p{2.2cm}|c|c|c|c|c|c|c|c|c|}
\hline
\multirow{2}{*}{Model} &
\multirow{2}{*}{Params} &
\multirow{2}{*}{\begin{tabular}[c]{@{}c@{}}Avg. \\ Training\\ Time\end{tabular}} &
\multirow{2}{*}{\begin{tabular}[c]{@{}c@{}}Avg. \\ Inference\\ Time\end{tabular}} &
\multicolumn{3}{c|}{\textbf{Category 1 Dataset}} & 
\multicolumn{3}{c|}{\textbf{Category 2 Dataset}} \\ \cline{5-10} 
                         &                          & & & \textbf{\begin{tabular}[c]{@{}c@{}}w/o OCR \\ Error\end{tabular}} & \textbf{\begin{tabular}[c]{@{}c@{}}w/ OCR \\ Error\end{tabular}} & \textbf{\begin{tabular}[c]{@{}c@{}}Perf. \\ Drop\end{tabular}} & \textbf{\begin{tabular}[c]{@{}c@{}}w/o OCR \\ Error\end{tabular}} & \textbf{\begin{tabular}[c]{@{}c@{}}w/ OCR \\ Error\end{tabular}} & \textbf{\begin{tabular}[c]{@{}c@{}}Perf. \\Drop\end{tabular}} \\ \hline
BERT\textsubscript{BASE}               & 110M  & 27 mins & 959 ms & 89.84 & 64.60 & 25.24 & 92.03 & 58.97 & 33.06 \\ \hline
Distill-BERT           & 65M   & {\ul25 mins} & \textbf{565 ms} & 90.50 & 59.12 & 31.38 & 93.63 & 55.71 & 37.91 \\ \hline
SDMGR                  & 5M    & 28 mins & 1207 ms & 89.14 & 87.03 & {\ul 2.11} & 98.03 & 94.65 & {\ul3.38} \\ \hline
LayoutLMv2\textsubscript{BASE}         & 200M  & 44 mins & 1907 ms & 94.03 & 74.57 & 19.46 & 93.26 & 89.71 & 3.55 \\ \hline
LayoutLMv3\textsubscript{BASE}         & 125M  & 35 mins & 1363 ms & {\ul 97.24} & {\ul 91.40} & 5.84 & {\ul 99.31} & {\ul 95.77} & 3.54 \\ \hline
\textbf{FS-DAG (ours)} & 81M   &  \textbf{21 mins} & {\ul773 ms} & \textbf{98.89} & \textbf{97.96} & \textbf{0.93} & \textbf{99.93} & \textbf{99.02} & \textbf{0.91} \\ \hline
\end{tabular}
\caption{Summary of model complexity, performance, robustness, and computational efficiency across five document types in the Category 1 dataset and seven document types in the Category 2 dataset. The best performance is highlighted in bold, and the second-best is underlined.}
\label{tab:table_2}
\end{table*}

\subsection{Results and Discussions}

We extensively conduct experiments with the two industrial dataset categories, owing to their diversity and industry relevance compared to publicly existing datasets. For benchmarking the models, we used five documents for training, while the remaining documents were used for testing. The split pattern was consistent across all the document types in both dataset categories. All the experiments for FS-DAG and other state-of-the-art models were run thrice, and the average results of the three runs are reported. We report the average F1 score across the document types in each dataset category. 

\textbf{Few-shot Key Information Extraction (KIE) Task.} Column "w/o OCR Error" of Category 1 \& Category 2 Datasets of Table \ref{tab:table_2} summarises the average F1-score results for both the dataset categories mentioned in Table \ref{tab:table_1} when the input OCR results of the document has no detection or recognition errors. It can be seen that FS-DAG outperforms its peer models with a high-performance gap. It can also be seen that LayoutLMv3 outperforms LayoutLMv2 while reducing the model complexity. LayoutLMv3 has very competitive results with FS-DAG but has higher model complexity. FS-DAG's performance can be attributed to the pre-trained models plugged in as feature extractors and position embeddings in the graph layer. It is also observed that the performance of FS-DAG and LayoutLMv3 are similar though the model complexity differs. FS-DAG outperforms SDMG-R by 9.75\% and 1.9\% for category 1 and  2 datasets, respectively. It highlights that the proposed changes over other graph models enable FS-DAG to have competitive performance with other larger multi-model models. The detailed experiment results are presented in Appendix \ref{sec:experiments}.

\textbf{Model Robustness}. KIE models often depend on OCR engines to extract text, which are then used as input. Despite improvements, OCR engines still produce errors, particularly with poor-quality documents. Some LMMs (e.g., Donut, LLaVa) incorporate OCR capabilities but suffer from similar limitations while significantly increasing model size beyond 500M parameters. We assess model robustness to OCR and misspelling errors by measuring performance drops due to misclassification. A robust model shows minimal performance decline, while models heavily reliant on text modality exhibit a more significant drop.

To evaluate robustness, we train models with ground-truth OCR data but introduce standard OCR errors with a probability of 0.1 during inference using nlpaug \citep{ma2019nlpaug} (details in Appendix \ref{sec:experiments}). The average F1-scores under these conditions are shown in Column "w/ OCR Error" of Table \ref{tab:table_2}, with the performance drop reported in Column "Perf. Drop".

FS-DAG demonstrates consistent robustness to OCR and misspelling errors with a performance drop of less than 1\%, enhancing its reliability for real-world applications. Notably, SDMG-R also shows a lower performance drop compared to other models, underscoring the advantage of graph-based models in effectively integrating a document's modalities, as opposed to transformer-based models that heavily rely on textual sequences and tokenization \citep{pattnayak2025tokenizationmattersimprovingzeroshot}.

\textbf{Model Complexity.} Table \ref{tab:table_2} also compares the model parameters, training and inference time across models. FS-DAG has  substantially higher parameters compared graph-based SDMG-R owing to the pluggable pre-trained backbones. However, FS-DAG has almost 60-40\% fewer parameters than other pre-trained transformer-based models like LayoutLMv2 or LayoutLMv3. LayoutLMv3 has competitive results with FS-DAG but with 64\% additional model parameters.

The "Avg. Training Time" is reported against both the dataset categories for all the models. 
SDMG-R requires longer training because it's trained from scratch, unlike other models that are only fine-tuned. Additionally, training time increases with model size.


\begin{table}[!th]
\small
\centering
\renewcommand{\arraystretch}{1.2} 
\setlength{\tabcolsep}{10pt} 
\begin{tabular}{|l|c|c|}
\hline
\multicolumn{1}{|c|}{Model} & Params & \begin{tabular}[c]{@{}c@{}}Avg. Perf. (\%)\\ (F1-Score)\end{tabular} \\ \hline
BERT\textsubscript{BASE}        & 110M & 82.80          \\ \hline
Distill-BERT    & 65M  & 80.70          \\ \hline
SDMG-R         & 5M   & 82.80          \\ \hline
LayoutLMv2\textsubscript{BASE}  & 200M & { 86.00}    \\ \hline
LayoutLMv3\textsubscript{BASE}  & 125M & {\ul 87.14} \\ \hline
FS-DAG         & 81M  & \textbf{93.90}          \\ \hline
\end{tabular}
\caption{Summary of the average F1-Score (\%) across the 25 classes in the WildReceipt dataset. The best performance is highlighted in bold, while the second-best performance is underlined.}
\label{tab:table_3}
\end{table}
The "Avg. Inference Time" is reported against both the dataset categories for all the models. DistilBERT demonstrates the lowest latency but also exhibits lower performance across the datasets. FS-DAG achieves low latency while maintaining higher performance. Meanwhile, LayoutLMv3 has a latency that is 76\% higher than FS-DAG, offering competitive performance but with reduced robustness. The lower model complexity reduces the cost of adopting the proposed model for the industry while outperforming other models.

\textbf{Wildreceipt KIE Task.} Table \ref{tab:table_3} shows the average F1-score on the publicly available dataset WildReceipt \cite{sun2021spatial}, which extracts key-value pairs (25 classes) from restaurant receipts from various restaurants. The results reported here take an average of all the 25 classes in the dataset compared to the 12 classes reported by Sun \textit{etal~}\cite{sun2021spatial}. The results show that FS-DAG outperforms the graph-based model by 11.1\% while outperforming the LayoutLMv2 by 7.9\% and LayoutLMv3 by 6.76\% . These results demonstrate that FS-DAG is not only effective for a few-shot setting for a given document type but can scale across datasets with multiple-document types given sufficient training data with lesser model complexity.


\textbf{Effect of Domain-Specific Language Model:} We swap the pre-trained language model backbone (Distill-BERT) of FS-DAG with domain-specific language models for some of the datasets. The results in Table \ref{tab:table_5} and \ref{tab:table_6} showcase that using a language model which is better adapted to the finance and medical domain enables FS-DAG to perform better than using a generic language model as a textual feature extractor. Thus, the proposed modular architecture design enables higher performance in domain-specific use cases.

\begin{table}[!th]
\small
\centering
\renewcommand{\arraystretch}{1.2} 
\setlength{\tabcolsep}{3.4pt} 
\begin{center}
\begin{tabular}{|c|c|c|c|}
\hline
\begin{tabular}[c]{@{}c@{}}Base \\ Architecture\end{tabular} &
\begin{tabular}[c]{@{}c@{}}Langauge \\ Model used\end{tabular} &
\begin{tabular}[c]{@{}c@{}}\# of Params\\ (FS-DAG)\end{tabular} &
\begin{tabular}[c]{@{}c@{}}Ecommerce \\ Invoice\end{tabular} \\ \hline
\multirow{3}{*}{{\begin{tabular}[c]{@{}c@{}}FS-DAG\\ (proposed)\end{tabular}}} & Distill-BERT     & 81M  & 95.1           \\ \cline{2-4} 
                                                                                      & BERT\textsubscript{BASE}         & 110M & 96.26          \\ \cline{2-4} 
                                                                                      & ProsusAI/finbert & 125M & \textbf{98.63} \\ \hline
\end{tabular}
\end{center}
\caption{Results of replacing DistilBERT in FS-DAG with BERT and finance-domain-specific models on the eCommerce Invoice.}
\label{tab:table_5}
\end{table}

\begin{table}[!th]
\small
\centering
\renewcommand{\arraystretch}{1.2} 
\setlength{\tabcolsep}{4pt} 
\begin{center}
\begin{tabular}{|c|c|c|c|}
\hline
\begin{tabular}[c]{@{}c@{}}Base \\ Architecture\end{tabular} &
\begin{tabular}[c]{@{}c@{}}Langauge \\ Model used\end{tabular} &
\begin{tabular}[c]{@{}c@{}}\# of Params\\ (FS-DAG)\end{tabular} &
\begin{tabular}[c]{@{}c@{}}Adverse \\ Reaction \\ Health Form\end{tabular} \\ \hline
\multirow{4}{*}{\begin{tabular}[c]{@{}c@{}}FS-DAG\\ (proposed)\end{tabular}} &
  Distill-BERT &
  81M &
  96.53 \\ \cline{2-4} 
 &
  BERT\textsubscript{BASE} &
  110M &
  97.13 \\ \cline{2-4} 
 &
  \begin{tabular}[c]{@{}c@{}}BiomedVLP-\\ CXR-BERT-\\ general\end{tabular} &
  125M &
  \textbf{98.98} \\ \hline
\end{tabular}
\end{center}
\caption{Results of replacing DistilBERT in FS-DAG with BERT and medical-domain-specific models on the medical form.}
\label{tab:table_6}
\end{table}
\begin{table*}[!th]
\small
\begin{center}
\begin{tabular}{|c|c|ccccc|c|c|}
\hline
\multirow{4}{*}{Model} &
  \multirow{4}{*}{\#} &
  \multicolumn{5}{c|}{Architectural Changes Proposed} &
  \multirow{4}{*}{\begin{tabular}[c]{@{}c@{}}Avg \\ Perf. (\%) \\ (F1 Score)\end{tabular}} &
  \multirow{4}{*}{\begin{tabular}[c]{@{}c@{}}Perf. \\ Gain (\%) \\ (F1 Score)\end{tabular}} \\ \cline{3-7}
 &
   &
  \multicolumn{1}{c|}{\begin{tabular}[c]{@{}c@{}}Pre-trained LM \\w/ first token \\ embedding\end{tabular}} &
  \multicolumn{1}{c|}{\begin{tabular}[c]{@{}c@{}}Pre-trained LM \\w/ pooling token \\ embeddings\end{tabular}} &
  \multicolumn{1}{c|}{\begin{tabular}[c]{@{}c@{}}Pre-trained \\ Visual Model\end{tabular}} &
  \multicolumn{1}{c|}{\begin{tabular}[c]{@{}c@{}}Position \\ Embedding \\ in GNN\end{tabular}} &
  \begin{tabular}[c]{@{}c@{}}Training \\ Strategies\end{tabular} &
   &
   \\ \hline
\multirow{9}{*}{FS-DAG} &
  1 &
  \multicolumn{1}{l|}{} &
  \multicolumn{1}{l|}{} &
  \multicolumn{1}{l|}{} &
  \multicolumn{1}{l|}{} &
  \multicolumn{1}{l|}{} &
  88.31 &
  NA \\ \cline{2-9} 
 &
  2a &
  \multicolumn{1}{c|}{\checkmark} &
  \multicolumn{1}{c|}{} &
  \multicolumn{1}{c|}{} &
  \multicolumn{1}{c|}{} &
   &
  89.26 &
  0.95 \\ 
 &
  2b &
  \multicolumn{1}{c|}{} &
  \multicolumn{1}{c|}{\checkmark} &
  \multicolumn{1}{c|}{} &
  \multicolumn{1}{c|}{} &
   &
  91.61 &
  3.30 \\ 
 &
  2c &
  \multicolumn{1}{c|}{} &
  \multicolumn{1}{c|}{} &
  \multicolumn{1}{c|}{\checkmark} &
  \multicolumn{1}{c|}{} &
   &
  91.33 &
  3.02 \\ 
 &
  2d &
  \multicolumn{1}{c|}{} &
  \multicolumn{1}{c|}{} &
  \multicolumn{1}{c|}{} &
  \multicolumn{1}{c|}{\checkmark} &
   &
  93.64 &
  5.33 \\ 
 &
  2e &
  \multicolumn{1}{c|}{} &
  \multicolumn{1}{c|}{} &
  \multicolumn{1}{c|}{} &
  \multicolumn{1}{c|}{} &
  \checkmark &
  93.86 &
  5.55 \\ \cline{2-9} 
 &
  3 &
  \multicolumn{1}{c|}{} &
  \multicolumn{1}{c|}{\checkmark} &
  \multicolumn{1}{c|}{\checkmark} &
  \multicolumn{1}{c|}{} &
   &
  92.43 &
  4.12 \\ \cline{2-9} 
 &
  4 &
  \multicolumn{1}{c|}{} &
  \multicolumn{1}{c|}{\checkmark} &
  \multicolumn{1}{c|}{\checkmark} &
  \multicolumn{1}{c|}{\checkmark} &
   &
  97.37 &
  9.06 \\ \cline{2-9} 
 &
  \textbf{5} &
  \multicolumn{1}{c|}{\textbf{}} &
  \multicolumn{1}{c|}{\textbf{\checkmark}} &
  \multicolumn{1}{c|}{\textbf{\checkmark}} &
  \multicolumn{1}{c|}{\textbf{\checkmark}} &
  \textbf{\checkmark} &
  \textbf{98.89} &
  \textbf{10.58} \\ \hline
\end{tabular}
\end{center}
\caption{The detailed ablation study results on different components and training of FS-DAG are reported for the Category 1 dataset. We observe that each proposed change has a significant positive impact on the model performance. The final proposed architecture of FS-DAG configuration is shown in experiment row \textbf{\#5}. }
\label{tab:table_4}
\end{table*}

\subsection{Ablation Study}

We performed an ablation study on the industrial dataset to evaluate the effects of the architectural and training modifications, as detailed in Table \ref{tab:table_4}. 
The starting point for each experiment is the skeleton FS-DAG architecture (row  \#1), with node and edge dimensions as 64. From rows \#2s to \#2e in Table \ref{tab:table_4}, we study the individual contribution of the proposed changes in the few-shot setting. The results show that each component individually leads to a performance gain between 2\%-6\%. From rows \#3 to \#5 in Table \ref{tab:table_4}, we combine the individual component and observe a performance gain increasing from 4\% to 10\% against row \#1. The experiments conclusively show the importance and impact of the proposed changes and training for FS-DAG.

\textbf{Effect of Pre-trained Language Model:} We use Distill-BERT as the pluggable pre-trained language model for all the experiments for extracting textual features. Adding a pre-trained language model and using the first sub-token to represent a text region $\{r_i\}$ improves the F1-score by 0.95\% on average (Table \ref{tab:table_4}: From \#1 vs. \#2a). Further pooling all the sub-token representations of a text region $\{r_i\}$ to get the token representation, we see the performance improves by 3.30\% on average (Table \ref{tab:table_4}: From \#1 vs. \#2b). It highlights that pooling the sub-token representation to represent a text region $\{r_i\}$ gives a better and richer representation that enables the model to learn in a few-shot setting.

\textbf{Effect of Pretrained Visual Model:} We use UNET with a Resnet-18 backbone pre-trained on PubTabnet \cite{smock2022pubtables} for extracting the visual features. The model F1-score increases by 3.02\% (Table \ref{tab:table_4}: From \#1 vs. \#2c) on average across the five few-shot datasets. It highlights that using a pre-trained visual feature extractor enables FS-DAG to learn better in a few-shot setting. However, it can also be seen that the impact of pre-trained visual features is lesser than the textual features.

\textbf{Effect of Position Embedding:} We introduce learnable position embedding in the GNN layer of the model. The model F1-score increases by 5.33\% (Table \ref{tab:table_4}: From \#1 vs. \#2d)  on average across the five datasets, showcasing that the position embedding plays an essential role in the GNN layers learning, helping it to adapt to the given document type.

\textbf{Effect of Training Strategies:} Apart from the model architecture changes, the training strategy for models in a few-shot learning environment plays an important role. The proposed training strategies for FS-DAG led to an increase of F1-score of 5.55\% (Table \ref{tab:table_4}: From \#1 vs. \#2e) on average across the five datasets. 

Finally, combining the different components shows an improvement (Table \ref{tab:table_4}: From \#3 to \#5), showcasing that the proposed components complement each other and leading to an overall average gain of 9.28\% for the proposed model in a few-shot setting.

\section{Conclusion}

FS-DAG presents a compelling alternative to large-scale models like VLMs, LMMs and LLMs, particularly for visually rich document understanding tasks in industrial applications like document classification, key value extraction, entity-linking and graph classification. By focusing on efficiency, scalability, and practical deployment, FS-DAG addresses the key limitations of these larger models, including their high computational cost and the challenges associated with training and running them in resource-constrained environments.

This work demonstrates FS-DAG's technical strengths and emphasizes its practical application in real-world environments, where its robustness, customizability, and low computational demands significantly lower operational costs, making advanced models more accessible across various industries. Currently, FS-DAG is adopted by over 50+ customers and provided through hyperscale cloud providers with over 1M+ API calls monthly.

Future research will focus on extending FS-DAG's capabilities to zero-shot learning and enhancing its adaptability to a broader range of industrial scenarios.




\bibliography{custom}

@inproceedings{rastogi2020information,
  title={Information Extraction From Document Images via FCA-Based Template Detection and Knowledge Graph Rule Induction},
  author={Rastogi, Mouli and Ali, Syed Afshan and Rawat, Mrinal and Vig, Lovekesh and Agarwal, Puneet and Shroff, Gautam and Srinivasan, Ashwin},
  booktitle={Proceedings of the IEEE/CVF Conference on Computer Vision and Pattern Recognition Workshops},
  pages={558--559},
  year={2020}
}

@article{yao2021one,
  title={One-shot Key Information Extraction from Document with Deep Partial Graph Matching},
  author={Yao, Minghong and Liu, Zhiguang and Wang, Liangwei and Li, Houqiang and Zhuang, Liansheng},
  journal={arXiv preprint arXiv:2109.13967},
  year={2021}
}

@article{devlin2018bert,
  title={Bert: Pre-training of deep bidirectional transformers for language understanding},
  author={Devlin, Jacob and Chang, Ming-Wei and Lee, Kenton and Toutanova, Kristina},
  journal={arXiv preprint arXiv:1810.04805},
  year={2018}
}

@article{sanh2019distilbert,
  title={DistilBERT, a distilled version of BERT: smaller, faster, cheaper and lighter},
  author={Sanh, Victor and Debut, Lysandre and Chaumond, Julien and Wolf, Thomas},
  journal={arXiv preprint arXiv:1910.01108},
  year={2019}
}

@article{lan2019albert,
  title={Albert: A lite bert for self-supervised learning of language representations},
  author={Lan, Zhenzhong and Chen, Mingda and Goodman, Sebastian and Gimpel, Kevin and Sharma, Piyush and Soricut, Radu},
  journal={arXiv preprint arXiv:1909.11942},
  year={2019}
}

@inproceedings{xu2020layoutlm,
  title={Layoutlm: Pre-training of text and layout for document image understanding},
  author={Xu, Yiheng and Li, Minghao and Cui, Lei and Huang, Shaohan and Wei, Furu and Zhou, Ming},
  booktitle={Proceedings of the 26th ACM SIGKDD International Conference on Knowledge Discovery \& Data Mining},
  pages={1192--1200},
  year={2020}
}

@article{xu2020layoutlmv2,
  title={Layoutlmv2: Multi-modal pre-training for visually-rich document understanding},
  author={Xu, Yang and Xu, Yiheng and Lv, Tengchao and Cui, Lei and Wei, Furu and Wang, Guoxin and Lu, Yijuan and Florencio, Dinei and Zhang, Cha and Che, Wanxiang and others},
  journal={arXiv preprint arXiv:2012.14740},
  year={2020}
}

@inproceedings{huang2022layoutlmv3,
  title={Layoutlmv3: Pre-training for document ai with unified text and image masking},
  author={Huang, Yupan and Lv, Tengchao and Cui, Lei and Lu, Yutong and Wei, Furu},
  booktitle={Proceedings of the 30th ACM International Conference on Multimedia},
  pages={4083--4091},
  year={2022}
}

@article{xu2021layoutxlm,
  title={Layoutxlm: Multimodal pre-training for multilingual visually-rich document understanding},
  author={Xu, Yiheng and Lv, Tengchao and Cui, Lei and Wang, Guoxin and Lu, Yijuan and Florencio, Dinei and Zhang, Cha and Wei, Furu},
  journal={arXiv preprint arXiv:2104.08836},
  year={2021}
}

@article{liu2019graph,
  title={Graph convolution for multimodal information extraction from visually rich documents},
  author={Liu, Xiaojing and Gao, Feiyu and Zhang, Qiong and Zhao, Huasha},
  journal={arXiv preprint arXiv:1903.11279},
  year={2019}
}

@article{sun2021spatial,
  title={Spatial dual-modality graph reasoning for key information extraction},
  author={Sun, Hongbin and Kuang, Zhanghui and Yue, Xiaoyu and Lin, Chenhao and Zhang, Wayne},
  journal={arXiv preprint arXiv:2103.14470},
  year={2021}
}

@article{lee2022formnet,
  title={Formnet: Structural encoding beyond sequential modeling in form document information extraction},
  author={Lee, Chen-Yu and Li, Chun-Liang and Dozat, Timothy and Perot, Vincent and Su, Guolong and Hua, Nan and Ainslie, Joshua and Wang, Renshen and Fujii, Yasuhisa and Pfister, Tomas},
  journal={arXiv preprint arXiv:2203.08411},
  year={2022}
}

@inproceedings{yu2021pick,
  title={PICK: processing key information extraction from documents using improved graph learning-convolutional networks},
  author={Yu, Wenwen and Lu, Ning and Qi, Xianbiao and Gong, Ping and Xiao, Rong},
  booktitle={2020 25th International Conference on Pattern Recognition (ICPR)},
  pages={4363--4370},
  year={2021},
  organization={IEEE}
}

@misc{hpanwar2019detectron2,
  author =       {Himanshu},
  title =        {Detectron2},
  howpublished = {\url{https://github.com/hpanwar08/detectron2}},
  year =         {2019}
}

@article{ulyanov2016instance,
  title={Instance normalization: The missing ingredient for fast stylization},
  author={Ulyanov, Dmitry and Vedaldi, Andrea and Lempitsky, Victor},
  journal={arXiv preprint arXiv:1607.08022},
  year={2016}
}

@inproceedings{li2021structext,
  title={Structext: Structured text understanding with multi-modal transformers},
  author={Li, Yulin and Qian, Yuxi and Yu, Yuechen and Qin, Xiameng and Zhang, Chengquan and Liu, Yan and Yao, Kun and Han, Junyu and Liu, Jingtuo and Ding, Errui},
  booktitle={Proceedings of the 29th ACM International Conference on Multimedia},
  pages={1912--1920},
  year={2021}
}

@inproceedings{huang2019icdar2019,
  title={Icdar2019 competition on scanned receipt ocr and information extraction},
  author={Huang, Zheng and Chen, Kai and He, Jianhua and Bai, Xiang and Karatzas, Dimosthenis and Lu, Shijian and Jawahar, CV},
  booktitle={2019 International Conference on Document Analysis and Recognition (ICDAR)},
  pages={1516--1520},
  year={2019},
  organization={IEEE}
}

@article{gu2021domain,
  title={Domain-specific language model pretraining for biomedical natural language processing},
  author={Gu, Yu and Tinn, Robert and Cheng, Hao and Lucas, Michael and Usuyama, Naoto and Liu, Xiaodong and Naumann, Tristan and Gao, Jianfeng and Poon, Hoifung},
  journal={ACM Transactions on Computing for Healthcare (HEALTH)},
  volume={3},
  number={1},
  pages={1--23},
  year={2021},
  publisher={ACM New York, NY}
}

@article{lee2020biobert,
  title={BioBERT: a pre-trained biomedical language representation model for biomedical text mining},
  author={Lee, Jinhyuk and Yoon, Wonjin and Kim, Sungdong and Kim, Donghyeon and Kim, Sunkyu and So, Chan Ho and Kang, Jaewoo},
  journal={Bioinformatics},
  volume={36},
  number={4},
  pages={1234--1240},
  year={2020},
  publisher={Oxford University Press}
}

@inproceedings{liu2021finbert,
  title={Finbert: A pre-trained financial language representation model for financial text mining},
  author={Liu, Zhuang and Huang, Degen and Huang, Kaiyu and Li, Zhuang and Zhao, Jun},
  booktitle={Proceedings of the twenty-ninth international conference on international joint conferences on artificial intelligence},
  pages={4513--4519},
  year={2021}
}

@article{dwivedi2021graph,
  title={Graph neural networks with learnable structural and positional representations},
  author={Dwivedi, Vijay Prakash and Luu, Anh Tuan and Laurent, Thomas and Bengio, Yoshua and Bresson, Xavier},
  journal={arXiv preprint arXiv:2110.07875},
  year={2021}
}

@misc{ma2019nlpaug,
  title={NLP Augmentation},
  author={Edward Ma},
  howpublished={https://github.com/makcedward/nlpaug},
  year={2019}
}

@inproceedings{zhong2019publaynet,
  title={Publaynet: largest dataset ever for document layout analysis},
  author={Zhong, Xu and Tang, Jianbin and Yepes, Antonio Jimeno},
  booktitle={2019 International Conference on Document Analysis and Recognition (ICDAR)},
  pages={1015--1022},
  year={2019},
  organization={IEEE}
}

@inproceedings{he2016deep,
  title={Deep residual learning for image recognition},
  author={He, Kaiming and Zhang, Xiangyu and Ren, Shaoqing and Sun, Jian},
  booktitle={Proceedings of the IEEE conference on computer vision and pattern recognition},
  pages={770--778},
  year={2016}
}

@inproceedings{ronneberger2015u,
  title={U-net: Convolutional networks for biomedical image segmentation},
  author={Ronneberger, Olaf and Fischer, Philipp and Brox, Thomas},
  booktitle={Medical Image Computing and Computer-Assisted Intervention--MICCAI 2015: 18th International Conference, Munich, Germany, October 5-9, 2015, Proceedings, Part III 18},
  pages={234--241},
  year={2015},
  organization={Springer}
}

@inproceedings{jaume2019,
    title = {FUNSD: A Dataset for Form Understanding in Noisy Scanned Documents},
    author = {Guillaume Jaume, Hazim Kemal Ekenel, Jean-Philippe Thiran},
    booktitle = {Accepted to ICDAR-OST},
    year = {2019}
}

@inproceedings{he2017mask,
  title={Mask r-cnn},
  author={He, Kaiming and Gkioxari, Georgia and Doll{\'a}r, Piotr and Girshick, Ross},
  booktitle={Proceedings of the IEEE international conference on computer vision},
  pages={2961--2969},
  year={2017}
}

@article{muller2019does,
  title={When does label smoothing help?},
  author={M{\"u}ller, Rafael and Kornblith, Simon and Hinton, Geoffrey E},
  journal={Advances in neural information processing systems},
  volume={32},
  year={2019}
}

@article{orfew,
  title={Few-Shot Learning for Structured Information Extraction From Form-Like Documents Using a Diff Algorithm},
  author={Or, Nerya and Urbach, Shlomo}
}

@article{wang2022towards,
  title={Towards Few-shot Entity Recognition in Document Images: A Label-aware Sequence-to-Sequence Framework},
  author={Wang, Zilong and Shang, Jingbo},
  journal={arXiv preprint arXiv:2204.05819},
  year={2022}
}

@article{gralinski2020kleister,
  title={Kleister: A novel task for information extraction involving long documents with complex layout},
  author={Grali{\'n}ski, Filip and Stanis{\l}awek, Tomasz and Wr{\'o}blewska, Anna and Lipi{\'n}ski, Dawid and Kaliska, Agnieszka and Rosalska, Paulina and Topolski, Bartosz and Biecek, Przemys{\l}aw},
  journal={arXiv preprint arXiv:2003.02356},
  year={2020}
}

@article{wang2022benchmark,
  title={A Benchmark for Structured Extractions from Complex Documents},
  author={Wang, Zilong and Zhou, Yichao and Wei, Wei and Lee, Chen-Yu and Tata, Sandeep},
  journal={arXiv preprint arXiv:2211.15421},
  year={2022}
}

@article{wu2020comprehensive,
  title={A comprehensive survey on graph neural networks},
  author={Wu, Zonghan and Pan, Shirui and Chen, Fengwen and Long, Guodong and Zhang, Chengqi and Philip, S Yu},
  journal={IEEE transactions on neural networks and learning systems},
  volume={32},
  number={1},
  pages={4--24},
  year={2020},
  publisher={IEEE}
}

@article{khemani2024review,
  title={A review of graph neural networks: concepts, architectures, techniques, challenges, datasets, applications, and future directions},
  author={Khemani, Bharti and Patil, Shruti and Kotecha, Ketan and Tanwar, Sudeep},
  journal={Journal of Big Data},
  volume={11},
  number={1},
  pages={18},
  year={2024},
  publisher={Springer}
}

@article{brown2020language,
  title={Language models are few-shot learners},
  author={Brown, Tom B},
  journal={arXiv preprint arXiv:2005.14165},
  year={2020}
}

@article{touvron2023llama,
  title={Llama: Open and efficient foundation language models},
  author={Touvron, Hugo and Lavril, Thibaut and Izacard, Gautier and Martinet, Xavier and Lachaux, Marie-Anne and Lacroix, Timoth{\'e}e and Rozi{\`e}re, Baptiste and Goyal, Naman and Hambro, Eric and Azhar, Faisal and others},
  journal={arXiv preprint arXiv:2302.13971},
  year={2023}
}

@article{jiang2024mixtral,
  title={Mixtral of experts},
  author={Jiang, Albert Q and Sablayrolles, Alexandre and Roux, Antoine and Mensch, Arthur and Savary, Blanche and Bamford, Chris and Chaplot, Devendra Singh and Casas, Diego de las and Hanna, Emma Bou and Bressand, Florian and others},
  journal={arXiv preprint arXiv:2401.04088},
  year={2024}
}

@article{cheng2024survey,
  title={A Survey on Deep Neural Network Pruning: Taxonomy, Comparison, Analysis, and Recommendations},
  author={Cheng, Hongrong and Zhang, Miao and Shi, Javen Qinfeng},
  journal={IEEE Transactions on Pattern Analysis and Machine Intelligence},
  year={2024},
  publisher={IEEE}
}

@article{hu2022lora,
  title={LoRA: Low-Rank Adaptation of Large Language Models},
  author={Hu, Edward J and Shen, Yelong and Wallis, Phillip and Allen-Zhu, Zeyuan and Li, Yuanzhi and Wang, Shean and Wang, Lu and Chen, Weizhu},
  journal={arXiv preprint arXiv:2106.09685},
  year={2022},
  url={https://arxiv.org/abs/2106.09685}
}

@inproceedings{rausch2021docparser,
  title={Docparser: Hierarchical document structure parsing from renderings},
  author={Rausch, Johannes and Martinez, Octavio and Bissig, Fabian and Zhang, Ce and Feuerriegel, Stefan},
  booktitle={Proceedings of the AAAI Conference on Artificial Intelligence},
  volume={35},
  number={5},
  pages={4328--4338},
  year={2021}
}

@inproceedings{chen2024internvl,
  title={Internvl: Scaling up vision foundation models and aligning for generic visual-linguistic tasks},
  author={Chen, Zhe and Wu, Jiannan and Wang, Wenhai and Su, Weijie and Chen, Guo and Xing, Sen and Zhong, Muyan and Zhang, Qinglong and Zhu, Xizhou and Lu, Lewei and others},
  booktitle={Proceedings of the IEEE/CVF Conference on Computer Vision and Pattern Recognition},
  pages={24185--24198},
  year={2024}
}

@article{li2024llava,
  title={Llava-med: Training a large language-and-vision assistant for biomedicine in one day},
  author={Li, Chunyuan and Wong, Cliff and Zhang, Sheng and Usuyama, Naoto and Liu, Haotian and Yang, Jianwei and Naumann, Tristan and Poon, Hoifung and Gao, Jianfeng},
  journal={Advances in Neural Information Processing Systems},
  volume={36},
  year={2024}
}

@article{abdin2024phi,
  title={Phi-3 technical report: A highly capable language model locally on your phone},
  author={Abdin, Marah and Jacobs, Sam Ade and Awan, Ammar Ahmad and Aneja, Jyoti and Awadallah, Ahmed and Awadalla, Hany and Bach, Nguyen and Bahree, Amit and Bakhtiari, Arash and Behl, Harkirat and others},
  journal={arXiv preprint arXiv:2404.14219},
  year={2024}
}

@article{zhang2024vision,
  title={Vision-language models for vision tasks: A survey},
  author={Zhang, Jingyi and Huang, Jiaxing and Jin, Sheng and Lu, Shijian},
  journal={IEEE Transactions on Pattern Analysis and Machine Intelligence},
  year={2024},
  publisher={IEEE}
}

@article{kaddour2023challenges,
  title={Challenges and applications of large language models},
  author={Kaddour, Jean and Harris, Joshua and Mozes, Maximilian and Bradley, Herbie and Raileanu, Roberta and McHardy, Robert},
  journal={arXiv preprint arXiv:2307.10169},
  year={2023}
}

@article{park2019cord,
  title={CORD: A Consolidated Receipt Dataset for Post-OCR Parsing},
  author={Park, Seunghyun and Shin, Seung and Lee, Bado and Lee, Junyeop and Surh, Jaeheung and Seo, Minjoon and Lee, Hwalsuk},
  booktitle={Document Intelligence Workshop at Neural Information Processing Systems},
  year={2019}
}

@article{agarwal2021evaluate,
  title={EVALUATE GENERALISATION \& ROBUSTNESS OF VISUAL FEATURES FROM IMAGES TO VIDEO},
  author={AGARWAL, AMIT},
  year={2021},
  month        = {December},
  doi          = {10.13140/RG.2.2.33887.53928},
  advisor      = {Nasib Ullah},
 journal = {ResearchGate},
  note         = {Available at \url{https://doi.org/10.13140/RG.2.2.33887.53928}}
}

@article{patel2024llm,
  title={LLM for Barcodes: Generating Diverse Synthetic Data for Identity Documents},
  author={Patel, Hitesh Laxmichand and Agarwal, Amit and Kumar, Bhargava and Gupta, Karan and Pattnayak, Priyaranjan},
  journal={arXiv preprint arXiv:2411.14962},
  year={2024}
}

@misc{agarwal2023pseudo,
  title={Pseudo labelling for key-value extraction from documents},
  author={Agarwal, Amit and Pachauri, Kulbhushan},
  year={2023},
  month=nov # "~21",
  publisher={Google Patents},
  note={US Patent 11,823,478}
}

@misc{agarwal2024techniques,
  title={Techniques for graph data structure augmentation},
  author={Agarwal, Amit and Pachauri, Kulbhushan and Zadeh, Iman and Qian, Jun},
  year={2024},
  month=may # "~21",
  publisher={Google Patents},
  note={US Patent 11,989,964}
}

@misc{pattnayak2025clinicalqa20multitask,
      title={Clinical QA 2.0: Multi-Task Learning for Answer Extraction and Categorization}, 
      author={Priyaranjan Pattnayak and Hitesh Laxmichand Patel and Amit Agarwal and Bhargava Kumar and Srikant Panda and Tejaswini Kumar},
      year={2025},
      eprint={2502.13108},
      archivePrefix={arXiv},
      primaryClass={cs.CL},
      url={https://arxiv.org/abs/2502.13108}, 
}

@article{agarwal2024mvtamperbench,
  title={MVTamperBench: Evaluating Robustness of Vision-Language Models},
  author={Agarwal, Amit and Panda, Srikant and Charles, Angeline and Kumar, Bhargava and Patel, Hitesh and Pattnayak, Priyanranjan and Rafi, Taki Hasan and Kumar, Tejaswini and Chae, Dong-Kyu},
  journal={arXiv preprint arXiv:2412.19794},
  year={2024}
}

@article{pattnayak2024survey,
  title={Survey of Large Multimodal Model Datasets, Application Categories and Taxonomy},
  author={Pattnayak, Priyaranjan and Patel, Hitesh Laxmichand and Kumar, Bhargava and Agarwal, Amit and Banerjee, Ishan and Panda, Srikant and Kumar, Tejaswini},
  journal={arXiv preprint arXiv:2412.17759},
  year={2024}
}

@article{agarwal2024enhancing,
  title={Enhancing Document AI Data Generation Through Graph-Based Synthetic Layouts},
  author={Agarwal, Amit and Patel, Hitesh and Pattnayak, Priyaranjan and Panda, Srikant and Kumar, Bhargava and Kumar, Tejaswini},
  journal={arXiv preprint arXiv:2412.03590},
  year={2024}
}

@article{yin2024continuous,
  title={Continuous spiking graph neural networks},
  author={Yin, Nan and Wan, Mengzhu and Shen, Li and Patel, Hitesh Laxmichand and Li, Baopu and Gu, Bin and Xiong, Huan},
  journal={arXiv preprint arXiv:2404.01897},
  year={2024}
}

@article{pattnayak9339review,
  title={REVIEW OF REFERENCE GENERATION METHODS IN LARGE LANGUAGE MODELS},
  author={Pattnayak, Priyaranjan and Agarwal, Amit and Kumar, Bhargava and Bangera, Yeshil and Panda, Srikant and Kumar, Tejaswini and Patel, Hitesh Laxmichand},
  journal={Journal ID},
  volume={9339},
  pages={1263}
}

@misc{pattnayak2025tokenizationmattersimprovingzeroshot,
      title={Tokenization Matters: Improving Zero-Shot NER for Indic Languages}, 
      author={Priyaranjan Pattnayak and Hitesh Laxmichand Patel and Amit Agarwal},
      year={2025},
      eprint={2504.16977},
      archivePrefix={arXiv},
      primaryClass={cs.CL},
      url={https://arxiv.org/abs/2504.16977}, 
}

@inproceedings{patel2025sweeval,
  title={SweEval: Do LLMs Really Swear? A Safety Benchmark for Testing Limits for Enterprise Use},
  author={Patel, Hitesh Laxmichand and Agarwal, Amit and Das, Arion and Kumar, Bhargava and Panda, Srikant and Pattnayak, Priyaranjan and Rafi, Taki Hasan and Kumar, Tejaswini and Chae, Dong-Kyu},
  booktitle={Proceedings of the 2025 Conference of the Nations of the Americas Chapter of the Association for Computational Linguistics: Human Language Technologies (Volume 3: Industry Track)},
  pages={558--582},
  year={2025}
}

@inproceedings{pattnayak2025hybrid,
  title={Hybrid AI for Responsive Multi-Turn Online Conversations with Novel Dynamic Routing and Feedback Adaptation},
  author={Pattnayak, Priyaranjan and Agarwal, Amit and Meghwani, Hansa and Patel, Hitesh Laxmichand and Panda, Srikant},
  booktitle={Proceedings of the 4th International Workshop on Knowledge-Augmented Methods for Natural Language Processing},
  pages={215--229},
  year={2025}
}

@misc{panda2025out,
  title={Out of distribution element detection for information extraction},
  author={Panda, Srikant and Agarwal, Amit and Nambirajan, Gouttham and Pachauri, Kulbhushan},
  year={2025},
  publisher={Google Patents},
  note={US Patent App. 18/347,983}
}

@misc{thomas2025model,
  title={MODEL AUGMENTATION FRAMEWORK FOR DOMAIN ASSISTED CONTINUAL LEARNING IN DEEP LEARNING},
  author={Thomas, Edwin and Agarwal, Amit and Jana, Sandeep and Pachauri, Kulbhushan},
  year={2025},
  note={US Patent App. 18/406,905}
}

@misc{agarwal2025techniques,
  title={Techniques of information extraction for selection marks},
  author={Agarwal, Amit and Panda, Srikant and Pachauri, Kulbhushan},
  year={2025},
  publisher={Google Patents},
  note={US Patent App. 18/240,343}
}

@misc{panda2025techniques,
  title={Techniques of information extraction for selection marks},
  author={Panda, Srikant and Agarwal, Amit and Pachauri, Kulbhushan},
  year={2025},
  publisher={Google Patents},
  note={US Patent App. 18/240,344}
}

@inproceedings{smock2022pubtables,
  title={PubTables-1M: Towards comprehensive table extraction from unstructured documents},
  author={Smock, Brandon and Pesala, Rohith and Abraham, Robin},
  booktitle={Proceedings of the IEEE/CVF Conference on Computer Vision and Pattern Recognition},
  pages={4634--4642},
  year={2022}
}

@inproceedings{patel-etal-2025-pcri,
    title = "{PCRI}: Measuring Context Robustness in Multimodal Models for Enterprise Applications",
    author = "Patel, Hitesh Laxmichand  and
      Agarwal, Amit  and
      Panda, Srikant  and
      Meghwani, Hansa  and
      Dua, Karan  and
      Li, Paul  and
      Sheng, Tao  and
      Ravi, Sujith  and
      Roth, Dan",
    editor = "Potdar, Saloni  and
      Rojas-Barahona, Lina  and
      Montella, Sebastien",
    booktitle = "Proceedings of the 2025 Conference on Empirical Methods in Natural Language Processing: Industry Track",
    month = nov,
    year = "2025",
    address = "Suzhou (China)",
    publisher = "Association for Computational Linguistics",
    url = "https://aclanthology.org/2025.emnlp-industry.14/",
    doi = "10.18653/v1/2025.emnlp-industry.14",
    pages = "195--214",
    ISBN = "979-8-89176-333-3",
}

@inproceedings{agarwal-etal-2025-rci,
    title = "{RCI}: A Score for Evaluating Global and Local Reasoning in Multimodal Benchmarks",
    author = "Agarwal, Amit  and
      Patel, Hitesh Laxmichand  and
      Panda, Srikant  and
      Meghwani, Hansa  and
      Singh, Jyotika  and
      Dua, Karan  and
      Li, Paul  and
      Sheng, Tao  and
      Ravi, Sujith  and
      Roth, Dan",
    editor = "Potdar, Saloni  and
      Rojas-Barahona, Lina  and
      Montella, Sebastien",
    booktitle = "Proceedings of the 2025 Conference on Empirical Methods in Natural Language Processing: Industry Track",
    month = nov,
    year = "2025",
    address = "Suzhou (China)",
    publisher = "Association for Computational Linguistics",
    url = "https://aclanthology.org/2025.emnlp-industry.10/",
    doi = "10.18653/v1/2025.emnlp-industry.10",
    pages = "138--157",
    ISBN = "979-8-89176-333-3",
}

@misc{agarwal2024synthetic,
  title={Synthetic document generation pipeline for training artificial intelligence models},
  author={Agarwal, Amit and Panda, Srikant and Pachauri, Kulbhushan},
  year={2024},
  publisher={Google Patents},
  note={US Patent App. 17/994,712}
}

@inproceedings{dua-etal-2025-flexdoc,
    title = "{F}lex{D}oc: Parameterized Sampling for Diverse Multilingual Synthetic Documents for Training Document Understanding Models",
    author = "Dua, Karan  and
      Patel, Hitesh Laxmichand  and
      Mittal, Puneet  and
      Gupta, Ranjeet  and
      Agarwal, Amit  and
      Pabolu, Praneet  and
      Panda, Srikant  and
      Meghwani, Hansa  and
      Horwood, Graham  and
      Shah, Fahad",
    editor = "Potdar, Saloni  and
      Rojas-Barahona, Lina  and
      Montella, Sebastien",
    booktitle = "Proceedings of the 2025 Conference on Empirical Methods in Natural Language Processing: Industry Track",
    month = nov,
    year = "2025",
    address = "Suzhou (China)",
    publisher = "Association for Computational Linguistics",
    url = "https://aclanthology.org/2025.emnlp-industry.105/",
    doi = "10.18653/v1/2025.emnlp-industry.105",
    pages = "1500--1521",
    ISBN = "979-8-89176-333-3",
}

@inproceedings{agarwal-etal-2025-aligning-llms,
    title = "Aligning {LLM}s for Multilingual Consistency in Enterprise Applications",
    author = "Agarwal, Amit  and
      Meghwani, Hansa  and
      Patel, Hitesh Laxmichand  and
      Sheng, Tao  and
      Ravi, Sujith  and
      Roth, Dan",
    editor = "Potdar, Saloni  and
      Rojas-Barahona, Lina  and
      Montella, Sebastien",
    booktitle = "Proceedings of the 2025 Conference on Empirical Methods in Natural Language Processing: Industry Track",
    month = nov,
    year = "2025",
    address = "Suzhou (China)",
    publisher = "Association for Computational Linguistics",
    url = "https://aclanthology.org/2025.emnlp-industry.9/",
    doi = "10.18653/v1/2025.emnlp-industry.9",
    pages = "117--137",
    ISBN = "979-8-89176-333-3",
}

@inproceedings{meghwani-etal-2025-hard,
    title = "Hard Negative Mining for Domain-Specific Retrieval in Enterprise Systems",
    author = "Meghwani, Hansa  and
      Agarwal, Amit  and
      Pattnayak, Priyaranjan  and
      Patel, Hitesh Laxmichand  and
      Panda, Srikant",
    editor = "Rehm, Georg  and
      Li, Yunyao",
    booktitle = "Proceedings of the 63rd Annual Meeting of the Association for Computational Linguistics (Volume 6: Industry Track)",
    month = jul,
    year = "2025",
    address = "Vienna, Austria",
    publisher = "Association for Computational Linguistics",
    url = "https://aclanthology.org/2025.acl-industry.72/",
    doi = "10.18653/v1/2025.acl-industry.72",
    pages = "1013--1026",
    ISBN = "979-8-89176-288-6",

}

@misc{agarwal2024domain,
  title={Domain adapting graph networks for visually rich documents},
  author={Agarwal, Amit and Panda, Srikant and Karmakar, Deepak and Pachauri, Kulbhushan},
  year={2024},
  publisher={Google Patents},
  note={US Patent App. 18/240,480}
}

\appendix

\section{Appendix}
\label{sec:appendix}

\subsection{Details of Model Architecture}
\label{sec:appendix_model}
\subsubsection{Text Embeddings}

Training language models from scratch are resource-intensive, time-consuming, and needs to generalize better in a few-shot learning environment \cite{agarwal-etal-2025-aligning-llms,patel2025sweeval}. Hence, we designed our architecture to have a pluggable language model. It enables choosing multi-lingual domain-specific language models like BioBERT\citep{lee2020biobert}, BiomedNLP-PubMedBERT \citep{gu2021domain}, FinBERT \citep{liu2021finbert}, for various use cases requiring fine-grained features, like in the medical, finance, or law domain, while also helping choose regional or multi-lingual language-based models. Standard use cases can rely on models like BERT \citep{devlin2018bert}, Distill-BERT \citep{sanh2019distilbert}, and Alberta \citep{lan2019albert} based on the performance and latency requirement of the model. 	

As shown in Figure \ref{fig:onecol}, a document image I is parsed via an OCR engine (word-level) to extract text regions $\{r_i\}$. Formally, for a document with a total number of words $L$ we have the i-th $(0<i \leq L)$ text region as the i-th word in the document. We then cluster and sort the $\{r_i\}$ to get a consistent reading sequence $\{s\}$ for the document, which later enables us to extract contextual text representation using a pre-trained language model. The reading sequence $\{s\}$ is the document's reading order to ensure consistent feature extraction during training and inference.

The reading sequence $\{s\}$ is then passed through a language model which tokenizes and decodes the sequence to return a sequence of token embedding, where $y_j\ {\in R}^{D_t}$  is the text-embedding for the token in $\{s\}$,  $D_t$  is the dimension of the text embedding. The language model tokenizes the words/text regions $\{r_i\}$ within $\{s\}$ into multiple tokens for which we get the text embedding $\{y_j\}$. Hence, we pool text embeddings of the tokens belonging to a particular $\{r_i\}$ to get the textual embedding of the document's original word/text region.
During the model training, the language model weights are frozen, and the extracted textual embedding of the words/text regions $\{r_i\}$ is projected over linear layers to adapt them as per the document type. Formally, for a sequence of length $L$, we have the i-th text embedding as: 
\begin{equation}										
t_i\mathrm{\ =\ }{MLP}_1\mathrm{(LangModelEmb(}r_i\mathrm{))}
\end{equation}
${MLP}_1$ is a learnable multi-layer perceptron that fine-tunes the textual embedding of a word/text region from the Language Model. The LangModelEmb layer clusters and sorts the text regions $\{r_i\}$ to create the reading sequence $\{s\}$ and extracts and pools the token embeddings $\{y_j\}$ to create the textual embedding of the given word/text regions $\{r_i\}$.


\subsubsection{Visual Embeddings}

Text in documents is designed to capture human attention based on the text's color, font size, texture, and appearance. Hence to extract the visual features \cite{agarwal2021evaluate,patel-etal-2025-pcri,agarwal-etal-2025-rci}, we use a UNET \citep{ronneberger2015u} with a Resnet-18 \citep{he2016deep} backbone as a visual feature extractor. The Resnet-18 backbone is pre-trained on the document's dataset \citep{zhong2019publaynet,hpanwar2019detectron2} 
and can be swapped with any other feature extractor based on the document type. Since visual features in VRDs are very extensive and document type dependent, we do not freeze weights of the visual backbone, letting it adapt in the few-shot setting during end-to-end training. 

As shown in Figure \ref{fig:onecol}, a document image I is passed through the pre-trained visual model to extract feature maps. The RoI Align layer \citep{sun2021spatial,he2017mask} extracts the visual embedding ${v_i}$ for every text region $\{r_i\}$ using the bounding box coordinates on the output feature maps of the visual model. 
\begin{equation}
\label{eq:example}
v_i=RoIAlign(VisFeatMap\left(I),r_i\right)
\end{equation}
The VisFeatMap layer extracts the visual feature map based on the feature extractor backbone used. The RoIAlign layer extracts $\{v_i\}$ , where $v_i\ {\in R}^{D_v}$  based on the $\{r_i\}$ bounding box co-ordinates, and $D_v$ is the dimension of the visual embedding.

\subsubsection{Node \& Edge Embeddings}

The graph nodes $\{n_i\}$ are initialized by fusing the textual features $\{t_i\}$ and visual features $\{v_i\}$ in the deep fusion block as shown in Figure \ref{fig:onecol}. The deep fusion block uses the Kronecker product as per \citep{sun2021spatial} and projects the result on linear layers as : 
\begin{equation}										
n_i={MLP}_2\ (t_i\ \otimes \ v_i)	
\end{equation}
$\otimes$ is the Kronecker product operation, while ${MLP}_2$  is a learnable multi-layer perceptron, where $n_i\ {\in R}^{D_n}$ and $D_n$ is the dimension of the visual embedding.

The spatial relation $\{s_{ij}\}$ between the two connecting nodes $\{n_i\}$  and $\{n_j\}$, where $0<i,j\le L$, is defined by calculating the relative distance between the nodes using the bounding box coordinates $<x_0, y_0, x_1, y_1>$ as described in \citep{sun2021spatial,patel2024llm,agarwal2024enhancing}.
The spatial relation ${s_{ij}}$ is normalized after passing it through linear projection layers to initialize the edge embedding ${e_{ij}^\prime}$  as follows: 
\begin{equation}										
e_{ij}^\prime=N_{l2}({MLP}_3\left(s_{ij}\right)) 
\end{equation}
${MLP}_3$  is a learnable multi-layer perceptron that transforms the spatial relation information ${s_{ij}}$ into $e_{ij}^\prime$, where $e_{ij}\in R^{D_e}$ and $D_e$ dimension of the edge embedding. $N_{l2}$ is the $l_2$ normalization operation. In the  GNN layer, $e_{ij}^\prime$ interacts with the node and position embeddings to refine the edge embedding and interaction between nodes using multi-head attention.

\subsubsection{Position Embeddings \& Multi-head Attention}

We divide the entire document in a K x K grid as shown in Figure \ref{fig:onecol}, and all the text regions $\{r_i\}$ in a particular grid, share the same positional embedding. The positional embedding enables the graph module to learn more about a node's absolute positioning and neighbors. The size of the grid K becomes a hyper-parameter that can be updated based on the document type. In our experiments, we found the value of K=25 to work consistently well across all the datasets. 

Given a text region $\{r_i\}$, with the bounding box coordinates as $<x_0, y_0, x_1, y_1>$, the individual horizontal and vertical position embedding are computed as:
\begin{equation}										
{Pos}_{hor}={PosEmb}_{hor}\left(x_0\right)\ |\ |\ {PosEmb}_{hor}\left(x_1\right)
\end{equation}
\begin{equation}										
{Pos}_{ver}={PosEmb}_{ver}\left(y_0\right)\ |\ |\ {PosEmb}_{ver}\left(y_1\right)
\end{equation}
We separately learn the horizontal and vertical positional embedding. Finally, the positional embedding $\{p_i\}$, where $p_{i}\in R^{D_p}$ for a given node concatenates the horizontal and vertical positional embeddings and passes it through a non-linear function TanH as suggested in \citep{dwivedi2021graph}.
\begin{equation}										
p_i=\ {TanH(Pos}_{hor}\ |\ |\ \ {Pos}_{ver}\ )
\end{equation}
The positional embedding is integrated and trained during the message propagation along the edges and multi-head attention. The different attention heads focus on the groups and segments within the nodes that strongly influence each other. The attention scores enable dynamic weighing of the edge connections to enable better node feature aggregation along various positional grids.  
\begin{equation}										
e_{ij}^h={MLP}_4(\ n_i\ \left|\ \right|\ p_i\ \left|\ \right|\ e_{ij}^\prime\ \left|\ \right|\ n_j\ \left|\ \right|\ p_j)
\end{equation}

\begin{equation}										
\textbf{e}_{\textbf{ij}}^\textbf{h}={MLP}_5(e_{ij}^h)
\label{eq:eq_9}
\end{equation}
We concatenate the node embeddings $\{n_i\}$ and $\{n_j\}$ with their corresponding positional embedding $\{p_i\}$ and $\{p_j\}$ before concatenating it with the initial edge embedding $e_{ij}^\prime$ between them. ${MLP}_4$ is a multi-layer perceptron that transforms the concatenated embeddings for each attention head.  $e_{ij}^h\ {\in R}^{D_{ne\ }{X\ D}_h\ X\ D_n}$, where $D_{ne\ }$ represents the number of edges in the graph, $D_{h\ }$ represents the number of heads in the network and $D_{n\ }$ represents the node embedding dimension. ${MLP}_5$ is a multi-layer perceptron that transforms $e_{ij}^h$ into a scaler for each of the edges, where $\textbf{e}_{\textbf{ij}}^\textbf{h}\ {\in R}^{D_{ne\ }{X\ D}_h\ X\ 1}$. Finally, we refine the node features $\{n_i\}$ of the graph module K times as follows :
\begin{equation}										
n_i^{k+1}=\ n_i^k+\ \sigma(N_{IN}({MLP}_6^k(\concat_{h}^{}\sum_{j\neq i}{\alpha_{ij}^{kh}\textbf{e}_{\textbf{ij}}^{\textbf{kh}})))}
\label{eq:eq_10}
\end{equation}
where $n_i^{k\ }{\in R}^{D_{n\ }}$ indicates the features of the ith graph node at time step k. $\alpha_{ij}^{kh}$ is normalized edge weight at time step k for a particular attention head. $e_{ij}^{kh}$  is the transformed concatenated representation of a particular attention ahead at time step k as described in Equation \ref{eq:eq_9}. ${MLP}_6^k$  is a linear transformation at time step k. $N_{IN}$ is the instance norm over the embeddings before passing it through $\sigma$, which is the non-linear activation function ReLU. $\alpha_{ij}^{kh}$ is the learnable normalized weights between nodes $i$ and $j$ for every attention head h at time step k. It is given by : 
\begin{equation}										
\alpha_{ij}^{kh}=\frac{\exp(\textbf{e}_{\textbf{ij}}^\textbf{h})}{\sum_{j\neq i}{\exp(\textbf{e}_{\textbf{ij}}^\textbf{h})}}
\label{eq:eq_11}
\end{equation}
\begin{figure*}[!th]
\begin{center}
   
   \frame{\includegraphics[width=1\linewidth]{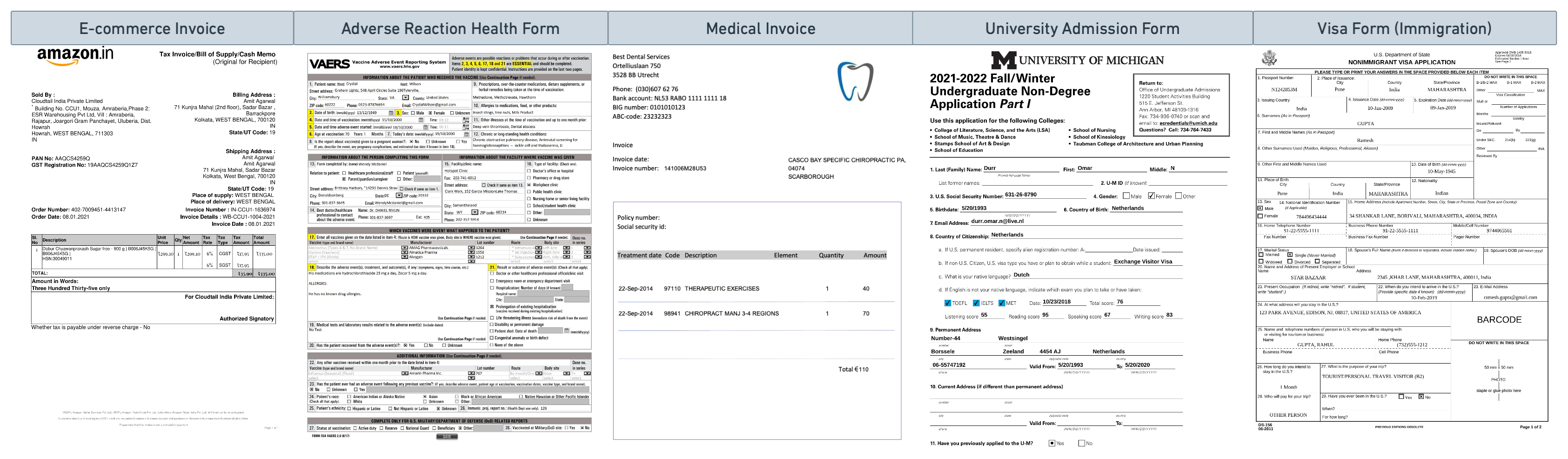}}
\end{center}
   \caption{Sample images from each of the five document types released as part of the Category 1 dataset.}
\label{fig:cat_1}
\end{figure*}

\begin{figure*}[!th]
\begin{center}
   
   \frame{\includegraphics[width=1\linewidth]{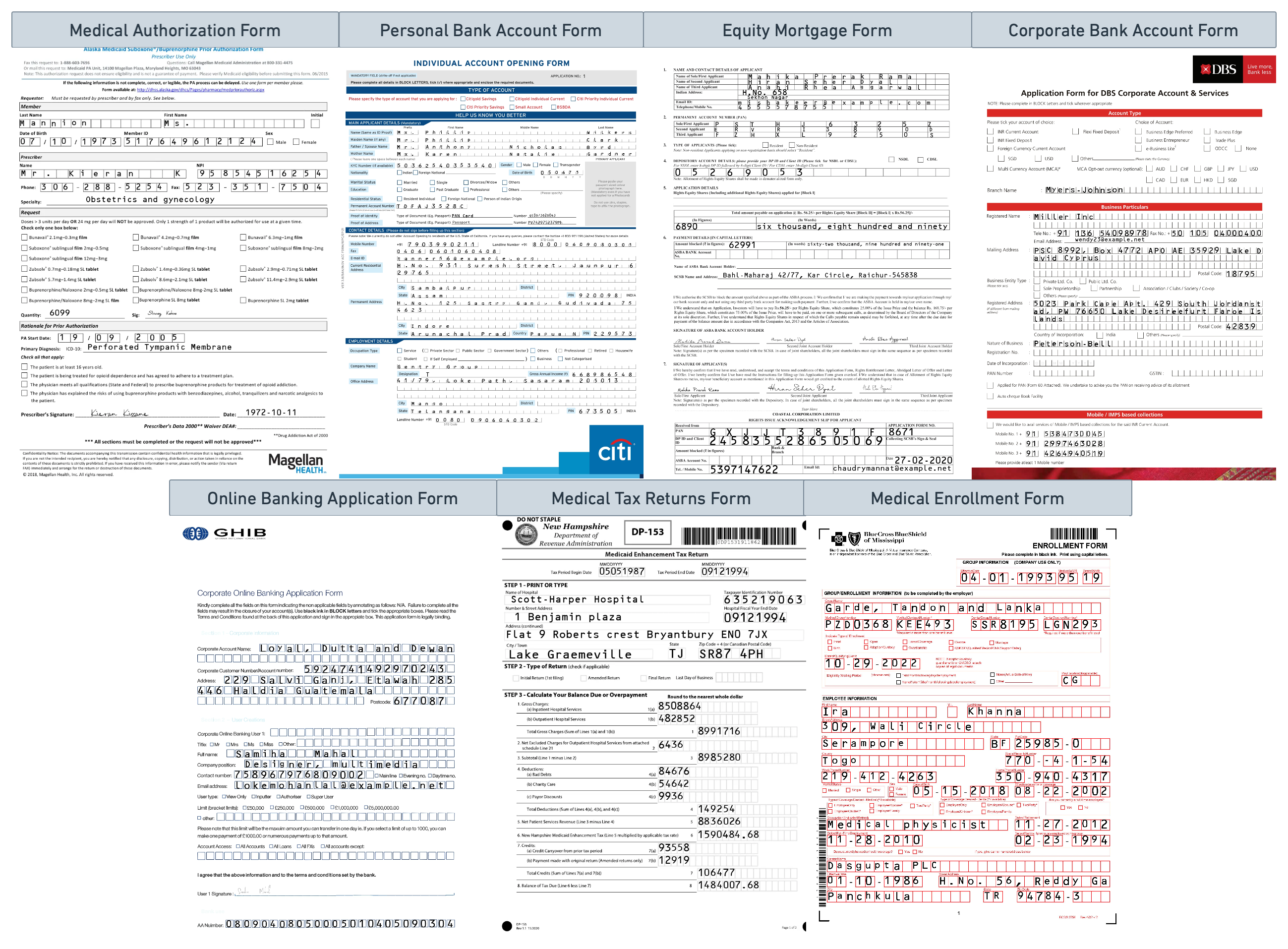}}
\end{center}
   \caption{Sample images from each of the seven document types released as part of the Category 2 dataset.}
\vspace{-1em}
\label{fig:cat_2}
\end{figure*}


\section{Experiments, Extended}
\label{sec:experiments}

\subsection{Dataset \& Metrics, Extended}

In Table \ref{tab:table_1} we share the class distribution of the various document types proposed in the dataset. Sample images for each document type \cite{agarwal2024synthetic,dua-etal-2025-flexdoc,meghwani-etal-2025-hard} in Category 1 are highlighted in Figure \ref{fig:cat_1}. In Figure \ref{fig:cat_2}, we highlight the sample images for each document type in Category 2. It can be seen that document types visually in Category 2 are fundamentally different from documents in Category 1 in how they are generated and filled with capturing necessary information for the business. These document types capture relevant information within specific placeholders, mostly filled character-by-character by a human or digital application. Document types in Category 2 datasets are still actively used worldwide, and more publicly available datasets for such documents must be available to steer research and evaluation of models. The released dataset will thus help further push boundaries for different document types in a few-shot setting.


\begin{table}
\begin{center}
\begin{tabular}{|cc|}
\hline
\multicolumn{1}{|c|}{\multirow{2}{*}{\textbf{Character}}} & \multirow{2}{*}{\textbf{Common OCR Errors}}                                       \\
\multicolumn{1}{|c|}{}                                    &                                                                                   \\ \hline
\multicolumn{1}{|c|}{\textbf{1}}                          & \begin{tabular}[c]{@{}c@{}}l(lower case of L),\\ I (Upper case of i)\end{tabular} \\ \hline
\multicolumn{1}{|c|}{\textbf{l (lowercase of L)}}         & I (Upper case of i)                                                               \\ \hline
\multicolumn{1}{|c|}{\textbf{6}}                          & b                                                                                 \\ \hline
\multicolumn{1}{|c|}{\textbf{5}}                          & S                                                                                 \\ \hline
\multicolumn{1}{|c|}{\textbf{,}}                          & .                                                                                 \\ \hline
\multicolumn{2}{|c|}{\textbf{Sample Augmentation}}                                                                                            \\ \hline
\multicolumn{1}{|c|}{\textbf{Original}}                   & \textbf{OCR Error Text}                                                           \\ \hline
\multicolumn{1}{|l|}{\begin{tabular}[c]{@{}l@{}}The quick brown fox\\  ate 5 chocolates\end{tabular}} &
  \multicolumn{1}{l|}{\begin{tabular}[c]{@{}l@{}}The quick brown fox\\  ate S chocoIates\end{tabular}} \\ \hline

\end{tabular}
\end{center}
\caption{Highlights most common OCR errors across popular OCR engines, along with a sample augmentation using nlpaug.}
\vspace{-1em} 
\label{tab:table_7}
\end{table}
\begin{table*}[th]
\small
\centering
\renewcommand{\arraystretch}{1.2} 
\setlength{\tabcolsep}{4pt} 
\begin{tabular}{|c|c|cccccc|}
\hline
\multirow{2}{*}{Model} &
  \multirow{2}{*}{Params} &
  \multicolumn{6}{c|}{F1- Score across Category 1 Dataset (Inference without OCR Errors)} \\ \cline{3-8} 
 &
   &
  \multicolumn{1}{c|}{\textbf{\begin{tabular}[c]{@{}c@{}}Ecommerce \\ Invoice\end{tabular}}} &
  \multicolumn{1}{c|}{\textbf{\begin{tabular}[c]{@{}c@{}}Adverse Reaction\\  Health Form\end{tabular}}} &
  \multicolumn{1}{c|}{\textbf{\begin{tabular}[c]{@{}c@{}}Medical \\ Invoice\end{tabular}}} &
  \multicolumn{1}{c|}{\textbf{\begin{tabular}[c]{@{}c@{}}University \\ Admission Form\end{tabular}}} &
  \multicolumn{1}{c|}{\textbf{\begin{tabular}[c]{@{}c@{}}Visa Form \\ (Immigration)\end{tabular}}} &
  \textbf{\begin{tabular}[c]{@{}c@{}}Avg  Perf.\end{tabular}} \\ \hline
BERT\textsubscript{BASE} &
  110M &
  \multicolumn{1}{c|}{91.60} &
  \multicolumn{1}{c|}{81.00} &
  \multicolumn{1}{c|}{98.60} &
  \multicolumn{1}{c|}{86.20} &
  \multicolumn{1}{c|}{91.80} &
  89.84 \\ \hline
Distill-BERT &
  65M &
  \multicolumn{1}{c|}{90.30} &
  \multicolumn{1}{c|}{82.50} &
  \multicolumn{1}{c|}{99.20} &
  \multicolumn{1}{c|}{90.70} &
  \multicolumn{1}{c|}{89.80} &
  90.50 \\ \hline
SDMGR &
  5M &
  \multicolumn{1}{c|}{90.58} &
  \multicolumn{1}{c|}{89.86} &
  \multicolumn{1}{c|}{90.15} &
  \multicolumn{1}{c|}{90.10} &
  \multicolumn{1}{c|}{85.01} &
  89.14 \\ \hline
LayoutLMv2\textsubscript{BASE} &
  200M &
  \multicolumn{1}{c|}{{\ul 97.20}} &
  \multicolumn{1}{c|}{88.60} &
  \multicolumn{1}{c|}{100.00} &
  \multicolumn{1}{c|}{95.97} &
  \multicolumn{1}{c|}{88.40} &
  94.03 \\ \hline
LayoutLMv3\textsubscript{BASE} &
  125M &
  \multicolumn{1}{c|}{ 95.80} &
  \multicolumn{1}{c|}{{\ul 95.00}} &
  \multicolumn{1}{c|}{\textbf{100.00}} &
  \multicolumn{1}{c|}{{\ul 97.20}} &
  \multicolumn{1}{c|}{{\ul 98.20}} &
  {\ul 97.24} \\ \hline
\textbf{FS-DAG (ours)} &
  81M &
  \multicolumn{1}{c|}{\textbf{98.30}} &
  \multicolumn{1}{c|}{\textbf{98.51}} &
  \multicolumn{1}{c|}{{\ul 99.90}} &
  \multicolumn{1}{c|}{\textbf{98.40}} &
  \multicolumn{1}{c|}{\textbf{99.34}} &
  \textbf{98.89} \\ \hline
\end{tabular}
\caption{Reports the field-level F1 scores for the KIE task in a few-shot learning setting for the five domain-specific document types from the category 1 dataset are reported. The best performance is highlighted in bold, while the second-best performance is underlined.}
\label{tab:table_8}
\end{table*}
\begin{table*}[!th]
\small
\centering
\renewcommand{\arraystretch}{1.2} 
\setlength{\tabcolsep}{4pt} 
\begin{tabular}{|c|cccccc|}
\hline
\multirow{2}{*}{Model} &
  \multicolumn{6}{c|}{F1 Score across Category 1 Dataset (Inference with OCR errors)} \\ \cline{2-7} 
 &
  \multicolumn{1}{c|}{\textbf{\begin{tabular}[c]{@{}c@{}}Ecommerce \\ Invoice\end{tabular}}} &
  \multicolumn{1}{c|}{\textbf{\begin{tabular}[c]{@{}c@{}}Adverse Reaction \\ Health Form\end{tabular}}} &
  \multicolumn{1}{c|}{\textbf{\begin{tabular}[c]{@{}c@{}}Medical Invoice\end{tabular}}} &
  \multicolumn{1}{c|}{\textbf{\begin{tabular}[c]{@{}c@{}}University \\ Admission Form\end{tabular}}} &
  \multicolumn{1}{c|}{\textbf{\begin{tabular}[c]{@{}c@{}}Visa Form \\ (Immigration)\end{tabular}}} &
  \textbf{\begin{tabular}[c]{@{}c@{}}Avg \\ Performance\end{tabular}} \\ \hline
BERT\textsubscript{BASE} &
  \multicolumn{1}{c|}{83.20} &
  \multicolumn{1}{c|}{36.30} &
  \multicolumn{1}{c|}{84.90} &
  \multicolumn{1}{c|}{60.40} &
  \multicolumn{1}{c|}{58.20} &
  64.60 \\ \hline
Distill-BERT &
  \multicolumn{1}{c|}{78.60} &
  \multicolumn{1}{c|}{38.70} &
  \multicolumn{1}{c|}{84.70} &
  \multicolumn{1}{c|}{46.30} &
  \multicolumn{1}{c|}{47.30} &
  59.12 \\ \hline
SDMGR &
  \multicolumn{1}{c|}{90.00} &
  \multicolumn{1}{c|}{{\ul 86.50}} &
  \multicolumn{1}{c|}{87.67} &
  \multicolumn{1}{c|}{87.00} &
  \multicolumn{1}{c|}{84.00} &
  87.03 \\ \hline
LayoutLMv2\textsubscript{BASE} &
  \multicolumn{1}{c|}{ 93.80} &
  \multicolumn{1}{c|}{42.30} &
  \multicolumn{1}{c|}{93.74} &
  \multicolumn{1}{c|}{85.00} &
  \multicolumn{1}{c|}{58.02} &
  74.57 \\ \hline
LayoutLMv3\textsubscript{BASE} &
  \multicolumn{1}{c|}{{{\ul95.40}}} &
  \multicolumn{1}{c|}{{81.20}} &
  \multicolumn{1}{c|}{{{\ul99.20}}} &
  \multicolumn{1}{c|}{{\ul 89.60}} &
  \multicolumn{1}{c|}{{\ul 91.60}} &
  {\ul 91.40} \\ \hline
{ \textbf{FS-DAG (ours)}} &
  \multicolumn{1}{c|}{\textbf{98.01}} &
  \multicolumn{1}{c|}{\textbf{97.93}} &
  \multicolumn{1}{c|}{\textbf{ 99.50}} &
  \multicolumn{1}{c|}{\textbf{96.80}} &
  \multicolumn{1}{c|}{\textbf{97.56}} &
  \textbf{97.96} \\ \hline
\end{tabular}
\caption{Reports the field-level F1 scores for the KIE tasks when the models are trained with ground-truth OCR (without any errors), and testing happens with words having OCR errors with a probability of 0.1.  FS-DAG outperforms the competitor models with a substantial performance gap, highlighting the generalizability and robustness of the model.The best performance is highlighted in bold, while the second-best performance is underlined.}
\label{tab:table_9}
\end{table*}
\begin{table*}[!htb]
\small
\centering
\renewcommand{\arraystretch}{1.2} 
\setlength{\tabcolsep}{4pt} 
\begin{tabular}{|c|cccccc|}
\hline
\multirow{2}{*}{Model} &
  \multicolumn{6}{c|}{Drop in F1 Score across Category 1 Dataset (Table \ref{tab:table_2} - Table \ref{tab:table_3})} \\ \cline{2-7} 
 &
  \multicolumn{1}{c|}{\textbf{\begin{tabular}[c]{@{}c@{}}Ecommerce \\ Invoice\end{tabular}}} &
  \multicolumn{1}{c|}{\textbf{\begin{tabular}[c]{@{}c@{}}Adverse Reaction \\ Health Form\end{tabular}}} &
  \multicolumn{1}{c|}{\textbf{\begin{tabular}[c]{@{}c@{}}Medical Invoice\end{tabular}}} &
  \multicolumn{1}{c|}{\textbf{\begin{tabular}[c]{@{}c@{}}University \\ Admission Form\end{tabular}}} &
  \multicolumn{1}{c|}{\textbf{\begin{tabular}[c]{@{}c@{}}Visa Form \\ (Immigration)\end{tabular}}} &
  \textbf{\begin{tabular}[c]{@{}c@{}}Avg \\ Perf. Drop\end{tabular}} \\ \hline
BERT\textsubscript{BASE} &
  \multicolumn{1}{c|}{8.40} &
  \multicolumn{1}{c|}{44.70} &
  \multicolumn{1}{c|}{13.70} &
  \multicolumn{1}{c|}{25.80} &
  \multicolumn{1}{c|}{33.60} &
  25.24 \\ \hline
Distill-BERT &
  \multicolumn{1}{c|}{11.70} &
  \multicolumn{1}{c|}{43.80} &
  \multicolumn{1}{c|}{14.50} &
  \multicolumn{1}{c|}{44.40} &
  \multicolumn{1}{c|}{42.50} &
  31.38 \\ \hline
SDMGR &
  \multicolumn{1}{c|}{0.58} &
  \multicolumn{1}{c|}{{\ul 3.36}} &
  \multicolumn{1}{c|}{2.48} &
  \multicolumn{1}{c|}{{\ul3.10}} &
  \multicolumn{1}{c|}{{\ul1.01}} &
  {\ul 2.11} \\ \hline
LayoutLMv2\textsubscript{BASE} &
  \multicolumn{1}{c|}{{3.40}} &
  \multicolumn{1}{c|}{46.30} &
  \multicolumn{1}{c|}{6.26} &
  \multicolumn{1}{c|}{10.97} &
  \multicolumn{1}{c|}{30.38} &
  19.46 \\ \hline
LayoutLMv3\textsubscript{BASE} &
  \multicolumn{1}{c|}{{{\ul0.40}}} &
  \multicolumn{1}{c|}{{13.80}} &
  \multicolumn{1}{c|}{{ {\ul0.80}}} &
  \multicolumn{1}{c|}{{7.60}} &
  \multicolumn{1}{c|}{{6.60}} &
  {5.84} \\ \hline
{\textbf{FS-DAG (ours)}} &
  \multicolumn{1}{c|}{\textbf{0.29}} &
  \multicolumn{1}{c|}{\textbf{0.58}} &
  \multicolumn{1}{c|}{\textbf{0.40}} &
  \multicolumn{1}{c|}{\textbf{1.6}} &
  \multicolumn{1}{c|}{\textbf{1.78}} &
  \textbf{0.93} \\ \hline
\end{tabular}
\caption{Highlights the fall in model performance (difference between results in Table \ref{tab:table_2} vs. Table \ref{tab:table_3}) when the test document has misspelling or OCR errors with a probability of 0.1. FS-DAG shows the minimum drop in performance overall and consistently higher performance compared to other models. The best performance is highlighted in bold, while the second-best performance is
underlined}
\label{tab:table_10}
\end{table*}


\subsection{Results, Extended}

The main paper reports average results across the different datasets for various state-of-the-art models. Here, we present the results on individual document types across both the dataset category for fine-grained analysis.

\textbf{Model Robustness.} To stimulate real-world misspelling or OCR errors in documents \cite{agarwal2024synthetic,agarwal2024techniques,agarwal2025techniques,panda2025out,panda2025techniques,patel2024llm}, we use nlpaug \cite{ma2019nlpaug} to introduce text recognition errors during the inference of models. Table \ref{tab:table_7} showcases the most common errors observed across various human misspellings and available OCR engines. The benchmarking of all the document types across the dataset categories when input errors are introduced during inference are detailed in Table \ref{tab:table_9} and \ref{tab:table_12}. Finally, we observe the drop in performance for individual document types across the two dataset categories in Table \ref{tab:table_10} and \ref{tab:table_13}. The observations are discussed in the following sections.

\textbf{Category 1 Dataset (KIE Task).} Table \ref{tab:table_8} shows the F1-score results of FS-DAG on the five industry document types from the category 1 dataset while comparing it to other state-of-the-art models. All the models are trained and tested in this benchmark with ground-truth annotations. We can observe that FS-DAG outperforms most of its peers by a considerable margin. At the same time, LayoutLMv3 has very similar performance compared to FS-DAG, and the best model varies based on the dataset with a small margin. In Table \ref{tab:table_9}, we report the F1-score when the training has been done with ground-truth OCR annotations. At the same time, during inference, misspelling and OCR errors are introduced at the word level with a probability of 0.1. Table \ref{tab:table_10} reports the drop in performance when the model is tested under the two different scenarios as represented in Table \ref{tab:table_8} and \ref{tab:table_9}. Models which are robust to input errors or less dependent on textual modality show a lower drop in performance.

It is observed that language models like BERT\textsubscript{BASE} and Distill-BERT have the maximum drop in performance as they rely entirely on textual modality. Multimodal model like LayoutLMv2 shows a higher performance drop than LayoutLMv3, suggesting that LayoutLMv2 is more dependent on the textual features. FS-DAG has the least fall in performance, followed by SDMG-R, implying better robustness to misspelling or OCR errors. The best-performing model for different document types vary and is highlighted in bold in Table \ref{tab:table_8}. However, FS-DAG outperforms its peers with the most consistent performance with lesser model complexity.

\textbf{ Category 2 Dataset (KIE Task).} Table \ref{tab:table_11} shows the F1-score results of FS-DAG on the seven industry document types from the category 2 dataset while comparing it to other state-of-the-art models. All the models are trained and tested in this benchmark with ground-truth OCR annotations. We can observe that FS-DAG outperforms most of its peers by a considerable margin, while LayoutLMv3 has a similar performance. In Table \ref{tab:table_12}, we report the F1-score when the training has been done with ground-truth annotations. At the same time, during inference, misspelling and OCR errors are introduced at the word level with a probability of 0.1. Table \ref{tab:table_13} reports the drop in performance when the model is tested under the two different scenarios as represented in Table \ref{tab:table_11} and \ref{tab:table_12}. SDMG-R, LayoutLM Series have performance drop in similar range which is higher compared to FS-DAG. The best-performing model for different document types vary and is highlighted in bold in Table \ref{tab:table_11}. FS-DAG outperforms its peers with the most consistent performance with lesser model complexity. 
It is observed that language models like BERT\textsubscript{BASE} and Distill-BERT have the maximum drop in performance (comparatively higher than document types in Category 1) as they rely entirely on textual features. 

\newpage
\begin{table*}[!th]
\small
\centering
\renewcommand{\arraystretch}{1.2} 
\setlength{\tabcolsep}{4pt} 
\begin{tabular}{|c|c|ccccccc|c|}
\hline
\multirow{2}{*}{Models} &
  \multirow{2}{*}{Params} &
  \multicolumn{7}{c|}{F1- Score across Category 2 Dataset (Inference without OCR Errors)} &
  \multirow{2}{*}{\begin{tabular}[c]{@{}c@{}}Avg\\  Perf.\end{tabular}} \\ \cline{3-9}
 &
   &
  \multicolumn{1}{c|}{\begin{tabular}[c]{@{}c@{}}Medical \\ Authorization\end{tabular}} &
  \multicolumn{1}{c|}{\begin{tabular}[c]{@{}c@{}}Personal \\ Bank \\ Account\end{tabular}} &
  \multicolumn{1}{c|}{\begin{tabular}[c]{@{}c@{}}Equity \\ Mortage\end{tabular}} &
  \multicolumn{1}{c|}{\begin{tabular}[c]{@{}c@{}}Corporate \\ Bank \\ Account\end{tabular}} &
  \multicolumn{1}{c|}{\begin{tabular}[c]{@{}c@{}}Online \\ Banking \\ Application\end{tabular}} &
  \multicolumn{1}{c|}{\begin{tabular}[c]{@{}c@{}}Medical \\ Tax \\ Returns\end{tabular}} &
  \begin{tabular}[c]{@{}c@{}}Medical \\ Insurance \\ Enrollment\end{tabular} &
   \\ \hline
BERT\textsubscript{BASE} &
  110M &
  \multicolumn{1}{c|}{96.1} &
  \multicolumn{1}{c|}{95.3} &
  \multicolumn{1}{c|}{87.4} &
  \multicolumn{1}{c|}{92.4} &
  \multicolumn{1}{c|}{89.2} &
  \multicolumn{1}{c|}{89.1} &
  94.7 &
  92.03 \\ \hline
Distill-BERT &
  65M &
  \multicolumn{1}{c|}{95.7} &
  \multicolumn{1}{c|}{97} &
  \multicolumn{1}{c|}{92.3} &
  \multicolumn{1}{c|}{92} &
  \multicolumn{1}{c|}{91.1} &
  \multicolumn{1}{c|}{90.2} &
  97.1 &
  93.63 \\ \hline
SDMGR &
  5M &
  \multicolumn{1}{c|}{95.67} &
  \multicolumn{1}{c|}{99.13} &
  \multicolumn{1}{c|}{95.67} &
  \multicolumn{1}{c|}{99.7} &
  \multicolumn{1}{c|}{98.3} &
  \multicolumn{1}{c|}{99} &
  {\ul 98.77} &
  98.03 \\ \hline
LayoutLMv2\textsubscript{BASE} &
  200M &
  \multicolumn{1}{c|}{96.9} &
  \multicolumn{1}{c|}{88.1} &
  \multicolumn{1}{c|}{94.1} &
  \multicolumn{1}{c|}{96.4} &
  \multicolumn{1}{c|}{87.5} &
  \multicolumn{1}{c|}{97.9} &
  91.9 &
  93.26 \\ \hline
{LayoutLMv3\textsubscript{BASE}} &
  125M &
  \multicolumn{1}{c|}{{\ul 96.9}} &
  \multicolumn{1}{c|}{{\ul 99.9}} &
  \multicolumn{1}{c|}{{\ul \textbf{100}}} &
  \multicolumn{1}{c|}{{\ul 99.9}} &
  \multicolumn{1}{c|}{100} &
  \multicolumn{1}{c|}{100} &
  98.5 &
  {\ul 99.31} \\ \hline
\textbf{FS-DAG} &
  81M &
  \multicolumn{1}{c|}{\textbf{100}} &
  \multicolumn{1}{c|}{\textbf{100}} &
  \multicolumn{1}{c|}{{\ul 99.9}} &
  \multicolumn{1}{c|}{\textbf{100}} &
  \multicolumn{1}{c|}{\textbf{100}} &
  \multicolumn{1}{c|}{\textbf{100}} &
  \textbf{99.6} &
  \textbf{99.93} \\ \hline
\end{tabular}
\caption{Reports the field-level F1 scores for the KIE task in a few-shot learning setting for the seven domain-specific document types from the category 2 dataset are reported. The best performance is highlighted in bold, while the second-best performance is underlined. }
\label{tab:table_11}
\end{table*}

\begin{table*}[!ht]
\small
\centering
\renewcommand{\arraystretch}{1.2} 
\setlength{\tabcolsep}{4pt} 
\begin{tabular}{|c|c|ccccccc|c|}
\hline
\multirow{2}{*}{Models} &
  \multirow{2}{*}{Params} &
  \multicolumn{7}{c|}{F1- Score across Category 2 Dataset(Inference with OCR errors)} &
  \multirow{2}{*}{\begin{tabular}[c]{@{}c@{}}Avg \\ Perf.\end{tabular}} \\ \cline{3-9}
 &
   &
  \multicolumn{1}{c|}{\begin{tabular}[c]{@{}c@{}}Medical \\ Authorization\end{tabular}} &
  \multicolumn{1}{c|}{\begin{tabular}[c]{@{}c@{}}Personal \\ Bank \\ Account\end{tabular}} &
  \multicolumn{1}{c|}{\begin{tabular}[c]{@{}c@{}}Equity \\ Mortage\end{tabular}} &
  \multicolumn{1}{c|}{\begin{tabular}[c]{@{}c@{}}Corporate \\ Bank \\ Account\end{tabular}} &
  \multicolumn{1}{c|}{\begin{tabular}[c]{@{}c@{}}Online \\ Banking \\ Application\end{tabular}} &
  \multicolumn{1}{c|}{\begin{tabular}[c]{@{}c@{}}Medical \\ Tax \\ Returns\end{tabular}} &
  \begin{tabular}[c]{@{}c@{}}Medical \\ Insurance \\ Enrollment\end{tabular} &
   \\ \hline
BERT\textsubscript{BASE} &
  110M &
  \multicolumn{1}{c|}{50.60} &
  \multicolumn{1}{c|}{40.80} &
  \multicolumn{1}{c|}{67.40} &
  \multicolumn{1}{c|}{58.90} &
  \multicolumn{1}{c|}{75.30} &
  \multicolumn{1}{c|}{69.00} &
  50.80 &
  58.97 \\ \hline
Distill-BERT &
  65M &
  \multicolumn{1}{c|}{40.30} &
  \multicolumn{1}{c|}{42.60} &
  \multicolumn{1}{c|}{64.90} &
  \multicolumn{1}{c|}{50.70} &
  \multicolumn{1}{c|}{77.70} &
  \multicolumn{1}{c|}{66.00} &
  47.80 &
  55.71 \\ \hline
SDMGR &
  5M &
  \multicolumn{1}{c|}{88.27} &
  \multicolumn{1}{c|}{90.70} &
  \multicolumn{1}{c|}{95.23} &
  \multicolumn{1}{c|}{98.37} &
  \multicolumn{1}{c|}{{\ul 99.10}} &
  \multicolumn{1}{c|}{98.47} &
  {\ul 92.40} &
  94.65 \\ \hline
LayoutLMv2\textsubscript{BASE} &
  200M &
  \multicolumn{1}{c|}{93.24} &
  \multicolumn{1}{c|}{80.19} &
  \multicolumn{1}{c|}{97.28} &
  \multicolumn{1}{c|}{91.39} &
  \multicolumn{1}{c|}{89.43} &
  \multicolumn{1}{c|}{91.12} &
  85.31 &
  89.71 \\ \hline
{LayoutLMv3\textsubscript{BASE}} &
  125M &
  \multicolumn{1}{c|}{{\ul 88.60}} &
  \multicolumn{1}{c|}{{\ul 98.00}} &
  \multicolumn{1}{c|}{{\textbf{99.45}}} &
  \multicolumn{1}{c|}{{\ul 95.37}} &
  \multicolumn{1}{c|}{{98.49}} &
  \multicolumn{1}{c|}{{ \ul 99.84}} &
  90.61 &
  {\ul 95.77} \\ \hline
\textbf{FS-DAG} &
  81M &
  \multicolumn{1}{c|}{\textbf{98.40}} &
  \multicolumn{1}{c|}{\textbf{98.50}} &
  \multicolumn{1}{c|}{{\ul 99.09}} &
  \multicolumn{1}{c|}{\textbf{99.43}} &
  \multicolumn{1}{c|}{\textbf{99.5}} &
  \multicolumn{1}{c|}{{ \textbf{99.67}}} &
  \textbf{96.57} &
  \textbf{99.02} \\ \hline
\end{tabular}

\caption{Reports the field-level F1 scores for the KIE tasks when the models are trained with ground-truth OCR (without any errors), and testing happens with words having OCR errors with a probability of 0.1.  FS-DAG outperforms the competitor models with a substantial performance gap, highlighting the generalizability and robustness of the model.The best performance is highlighted in bold, while the second-best performance is underlined.}
\label{tab:table_12}
\end{table*}
\begin{table*}[!ht]
\small
\centering
\renewcommand{\arraystretch}{1.2} 
\setlength{\tabcolsep}{4pt} 
\begin{tabular}{|c|c|ccccccc|c|}
\hline
\multirow{2}{*}{Models} &
  \multirow{2}{*}{Params} &
  \multicolumn{7}{c|}{Drop in F1 Score across Category 2 Dataset (Table \ref{tab:table_5} - \ref{tab:table_6})} &
  \multirow{2}{*}{\begin{tabular}[c]{@{}c@{}}Avg \\Perf.\\ Drop\end{tabular}} \\ \cline{3-9}
 &
   &
  \multicolumn{1}{c|}{\begin{tabular}[c]{@{}c@{}}Medical \\ Authorization\end{tabular}} &
  \multicolumn{1}{c|}{\begin{tabular}[c]{@{}c@{}}Personal \\ Bank \\ Account\end{tabular}} &
  \multicolumn{1}{c|}{\begin{tabular}[c]{@{}c@{}}Equity \\ Mortgage\end{tabular}} &
  \multicolumn{1}{c|}{\begin{tabular}[c]{@{}c@{}}Corporate \\ Bank \\ Account\end{tabular}} &
  \multicolumn{1}{c|}{\begin{tabular}[c]{@{}c@{}}Online \\ Banking \\ Application\end{tabular}} &
  \multicolumn{1}{c|}{\begin{tabular}[c]{@{}c@{}}Medical \\ Tax \\ Returns\end{tabular}} &
  \begin{tabular}[c]{@{}c@{}}Medical \\ Insurance \\ Enrollment\end{tabular} &
   \\ \hline
BERT\textsubscript{BASE} &
  110M &
  \multicolumn{1}{c|}{45.50} &
  \multicolumn{1}{c|}{54.50} &
  \multicolumn{1}{c|}{20.00} &
  \multicolumn{1}{c|}{33.50} &
  \multicolumn{1}{c|}{13.90} &
  \multicolumn{1}{c|}{20.10} &
  43.90 &
  33.06 \\ \hline
Distill-BERT &
  65M &
  \multicolumn{1}{c|}{55.40} &
  \multicolumn{1}{c|}{54.40} &
  \multicolumn{1}{c|}{27.40} &
  \multicolumn{1}{c|}{41.30} &
  \multicolumn{1}{c|}{13.40} &
  \multicolumn{1}{c|}{24.20} &
  49.30 &
  37.91 \\ \hline
SDMGR &
  5M &
  \multicolumn{1}{c|}{7.40} &
  \multicolumn{1}{c|}{8.43} &
  \multicolumn{1}{c|}{{\textbf{ 0.44}}} &
  \multicolumn{1}{c|}{{\ul 1.33}} &
  \multicolumn{1}{c|}{{\ul 0.80}} &
  \multicolumn{1}{c|}{0.53} &
  {\ul 6.37} &
  3.39 \\ \hline
LayoutLMv2\textsubscript{BASE} &
  200M &
  \multicolumn{1}{c|}{{\ul 3.66}} &
  \multicolumn{1}{c|}{7.91} &
  \multicolumn{1}{c|}{{3.18}} &
  \multicolumn{1}{c|}{5.01} &
  \multicolumn{1}{c|}{{1.93}} &
  \multicolumn{1}{c|}{6.78} &
  6.59 &
  3.55 \\ \hline
LayoutLMv3\textsubscript{BASE} &
  125M &
  \multicolumn{1}{c|}{8.30} &
  \multicolumn{1}{c|}{{\ul 1.90}} &
  \multicolumn{1}{c|}{{\ul 0.55}} &
  \multicolumn{1}{c|}{4.53} &
  \multicolumn{1}{c|}{1.51} &
  \multicolumn{1}{c|}{\textbf{0.16}} &
  7.89 &
  {\ul 3.55} \\ \hline
\textbf{FS-DAG} &
  81M &
  \multicolumn{1}{c|}{\textbf{1.60}} &
  \multicolumn{1}{c|}{\textbf{1.50}} &
  \multicolumn{1}{c|}{0.81} &
  \multicolumn{1}{c|}{\textbf{0.57}} &
  \multicolumn{1}{c|}{\textbf{0.50}} &
  \multicolumn{1}{c|}{{\ul 0.33}} &
  \textbf{1.03} &
  \textbf{0.91} \\ \hline
\end{tabular}
\caption{Highlights the fall in model performance (difference between results in Table \ref{tab:table_5} vs. Table \ref{tab:table_6}) when the test document has misspelling or OCR errors with a probability of 0.1. FS-DAG shows the minimum drop in performance overall and consistently higher performance compared to other models.}
\label{tab:table_13}
\end{table*}

\end{document}